\documentclass[a4paper,fleqn]{cas-sc}

\usepackage[square, comma, sort&compress, numbers]{natbib}
\usepackage{amssymb}
\usepackage{amsmath}

\usepackage{cases}
\usepackage{float} 
\usepackage{subfig}
\usepackage[ruled]{algorithm2e} 
\usepackage{diagbox}
\usepackage{verbatim}

\usepackage{multirow}

\usepackage[commandnameprefix=always]{changes}

\begin{document}




\title [mode = title]{Multi-Scale Separable Fourier Neural Networks for Solving High-Frequency PDEs} 
\shorttitle{MS-SFNN}

\author[1]{Qihong Yang}[style=chinese, orcid=0000-0002-8398-7212]
\ead{yangqh0808@163.com}
\address[1]{organization={School of Mathematics, Sichuan University},
    city={Chengdu},
    postcode={610065},
    country={China}}

\author[1]{Qiaolin He}[style=chinese]
\cormark[1]
\ead{qlhejenny@scu.edu.cn}

\begin{abstract}
Solving high-frequency partial differential equations (PDEs) with neural networks is notoriously difficult due to the spectral bias of conventional architectures. We propose the Multi-Scale Separable Fourier Neural Network (MS-SFNN), a framework designed to overcome this limitation by explicitly encoding multi-scale Fourier features within a separable representation. The network factorizes the solution into $d$ single-coordinate subnetworks with fixed, randomly initialized weights; these subnetworks are combined via element-wise products to form a rich set of basis functions. This separable construction scales linearly with the problem dimension, thus inherently alleviating the curse of dimensionality. Crucially, each subnetwork is equipped with trainable scaling factors coupled with cosine activations, providing an adaptive mechanism for multi-scale frequency selection that endows the model with strong spectral approximation capability. The PDE solution is expressed as a linear combination of these learned basis functions, and the combination coefficients are determined by solving a large-scale least-squares system. To resolve the memory bottleneck in high-frequency or three-dimensional settings, we replace automatic differentiation (AD) with analytical derivatives of the basis functions and use a memory-efficient batched QR decomposition for solving the least-squares systems efficiently. Extensive numerical experiments demonstrate that MS-SFNN achieves superior accuracy and substantially outperforms state-of-the-art methods, including Physics-Informed Neural Network (PINN) and the Separated-Variable Spectral Neural Network (SV-SNN).
\end{abstract}





\maketitle

\section{Introduction}
\label{sec:introduction}
In recent years, neural networks have attracted considerable interest in scientific computing, especially following the introduction of physics-informed neural networks (PINNs) \cite{PINN}. In contrast to classical grid-dependent methods such as the finite element method (FEM) \cite{ishihara1977convergence,ishihara1978mixed} and the finite difference method (FDM) \cite{truhlar1972finite,simos1997finite}, PINNs offer a mesh-free paradigm that trains directly on scattered collocation points. The network parameters are learned by minimizing a composite loss function encoding the partial differential equation (PDE) residual along with boundary and initial conditions, where differential operators are evaluated via automatic differentiation (AD), and optimizers such as Adam \cite{kingma2014adam} or L-BFGS \cite{liu1989limited} are commonly employed. This unified framework has been successfully applied to a broad spectrum of PDEs \cite{lu2021deepxde,pang2019fpinns,zhang2019quantifying,zhang2020learning} across optics \cite{chen2020physics,lu2021physics}, fluid mechanics \cite{raissi2020hidden}, systems biology \cite{yazdani2020systems}, and biomedicine \cite{sahli2020physics}. Despite this versatility, PINNs frequently yield lower accuracy than classical solvers such as FEM and FDM. This accuracy gap becomes particularly severe when the solution contains high-frequency components---a regime in which standard neural networks exhibit a well-known spectral bias towards low frequencies, severely limiting their capacity to resolve fine-scale features.

Efforts to improve the accuracy of PINNs have concentrated on five complementary strategies. First, the architecture space has expanded far beyond the standard multilayer perceptron, encompassing convolutional (CNNs) \cite{gao2021phygeonet,fang2021high,wandel2022spline}, recurrent \cite{ren2022phycrnet,mavi2023unsupervised}, generative adversarial (GANs) \cite{yang2020physics,gao2022wasserstein}, Kolmogorov – Arnold (KANs) \cite{wang2025kolmogorov}, binary structured (BsNN) \cite{BsPINN}, and Transformer \cite{dos2023physics,zhao2023pinnsformer} networks. Second, adaptive collocation point sampling \cite{lu2021deepxde,nabian2021efficient,WU2023115671,TANG2023111868} dynamically enriches points in high-residual regions, greatly improving resolution of steep gradients and rapid oscillations. Third, adaptive loss-weighting schemes \cite{WANG20M1318043,WANG2022110768,XIANG202211} counter the severe magnitude imbalance among loss terms. Fourth, input coordinate embeddings \cite{LI202460,LI2024113012,GUAN2023112360} lift the input into frequency-rich spaces to explicitly mitigate spectral bias. Fifth, domain decomposition strategies such as XPINN \cite{XPINN} and FBPINN \cite{FBPINN} partition the domain into independently trained sub-networks, extending PINNs to larger and more complex configurations. None of these modifications, however, have enabled PINNs to consistently match the accuracy and efficiency of classical solvers. Multi-stage training \cite{wang2024multi,aldirany2024multi} has recently attained high precision through sequential refinement, but at the cost of a several-fold increase in training time, severely hindering practical deployment

Randomized neural networks (RNNs) \cite{dong2022computing,dong2021local,chen2024optimization,chen2023random,RFM,shang2023randomized,shang2024randomized,wang2024randomized} form another prominent class of neural PDE solvers that recast the problem as a linear least-squares fit. The core idea is to use a randomly initialized network with frozen weights as a fixed functional basis; the PDE solution is then a linear combination of these basis functions, and the unknown coefficients are recovered by solving a well-conditioned linear system---completely bypassing the non-convex optimization that plagues standard PINNs. This design eliminates backpropagation entirely, yielding a deterministic and robust solver. The single-hidden-layer feedforward network (SLFN) is the predominant architecture in this paradigm, favored for its simplicity, universal approximation property, and low memory overhead when paired with AD. Nonetheless, an SLFN may lack the expressive capacity to span complex solution spaces. To address this, more expressive architectures such as HLConcELM \cite{ni2023numerical} have been proposed, boosting representational power through structured hierarchical concatenation. Overall, RNNs offer a compelling alternative to fully trained PINNs, particularly when accuracy, stability, and computational efficiency are the primary objectives.

In this work, we address the persistent difficulty of solving high-frequency PDEs with neural networks. Standard PINNs consistently fail to capture oscillatory behavior, and even enhanced variants \cite{zheng2025fg,app14083204,XiongAFFN,fang2024solving} that can qualitatively recover such solutions still fall short of achieving high accuracy. 
Notable efforts include FourierPINN \cite{FourierPINN}, which augments the network with tensor-product Fourier bases, and SR-HFNN/NR-HFNN \cite{SR-HFNN}, which approximate oscillatory solutions by constructing a dictionary of oscillation functions to form a linear combination of neural networks. Alongside SV-SNN \cite{SV-SNN}, a separated-variable spectral neural network, and LbNM \cite{LI2024200}, which utilizes Tikhonov regularization to stably learn the solution operator by leveraging relevant information---especially fundamental solutions, these methods have improved high-frequency learning, yet the accuracy gap remains. To overcome this limitation, we propose the Multi-Scale Separable Fourier Neural Network (MS-SFNN), an architecture purpose-built for high-frequency PDEs. MS-SFNN employs a separable representation: for a $d$-dimensional input, $d$ independent single-coordinate subnetworks with randomly initialized and frozen weights are combined via element-wise products to form basis functions. Each subnetwork incorporates a trainable scaling factor coupled with a cosine activation, explicitly embedding multi-scale Fourier features and enabling dimension-adaptive frequency modulation. Following the randomized neural network paradigm, the PDE solution is a linear combination of these separable bases, with coefficients determined by solving a linear least-squares problem. To relieve the severe memory pressure caused by dense collocation grids in high-frequency and three-dimensional settings, we forgo AD and instead employ analytical derivatives of the basis functions, substantially cutting memory usage. Furthermore, we use a batched QR decomposition that processes collocation points in chunks, providing a stable and memory-efficient solver for large-scale least-squares systems.

The remainder of this paper is organized as follows. Section \ref{sec:problems} briefly reviews the PINN and RNN methodologies for solving PDEs. Section \ref{sec:methods} describes the proposed MS-SFNN architecture. Section \ref{sec:experiments} reports a series of numerical experiments on high-frequency PDEs to demonstrate the accuracy, efficiency, and robustness of our approach. Section \ref{sec:conclusions} concludes the article with a summary of contributions and an outlook on future research directions.

\section{Preliminaries}
\label{sec:problems}

\subsection{Physics-Informed Neural Networks}
In this work, we consider PDEs of the form
\begin{equation}
    \label{eq:linear_PDEs}
    \begin{array}{r@{}l}
        \left\{
        \begin{aligned}
            \mathcal{L}u(\boldsymbol{x}) &= f(\boldsymbol{x}), &  & \mbox{in} \enspace \Omega,          \\
            \mathcal{B}u(\boldsymbol{x}) &= g(\boldsymbol{x}),        &  & \mbox{on} \enspace \partial \Omega, \\
        \end{aligned}
        \right.
    \end{array}
\end{equation}
where $u : \Omega \to \mathbb{R}$ is the unknown scalar field, $\boldsymbol{x} \in \mathbb{R}^d$ denotes the spatial (or space-time) coordinate, $\mathcal{L}$ is a differential operator defined on $\Omega \subset \mathbb{R}^d$, and $\mathcal{B}$ is a boundary operator acting on $\partial\Omega$. The functions $f$ and $g$ correspond to the source term and the prescribed boundary condition, respectively.

Raissi et al. \cite{PINN} introduced PINNs as a flexible framework for solving PDEs with neural networks, stimulating widespread research interest. Let $S_{e} = \{\boldsymbol{x}_i\}_{i=1}^{N_{e}}$ and $S_{b}=\{\boldsymbol{x}_i\}_{i=1}^{N_{b}}$ denote the sets of collocation points for the governing equation and boundary conditions, respectively, where the points may be uniformly spaced or randomly sampled within $\Omega$ and on $\partial\Omega$. 
Substituting these collocation points into the PDE system \eqref{eq:linear_PDEs} yields the following residual-based loss terms:
\begin{equation}
    \label{eq:PINN_loss}
    \begin{array}{r@{}l}
        \left\{
        \begin{aligned}
            Loss_{e} &= \frac{1}{N_e} \sum_{i=1}^{N_e} [\mathcal{L}u_{NN}(\boldsymbol{x}_i, t_i) - f(\boldsymbol{x}_i, t_i)]^2,   \\
            Loss_{b} & = \frac{1}{N_b} \sum_{i=1}^{N_b} [\mathcal{B}u_{NN}(\boldsymbol{x}_i, t_i) - g(\boldsymbol{x}_i, t_i)]^2, \\
        \end{aligned}
        \right.
    \end{array}
\end{equation}
where $u_{NN}=\mathcal{N}(\boldsymbol{x}; \boldsymbol{\Theta})$ is the neural network approximation parameterized by $\boldsymbol{\Theta}$. The total loss is formed as $Loss_{total} = \lambda_e Loss_e + \lambda_b Loss_b$, where $\lambda_e, \lambda_b > 0$ being positive weighting coefficients that balance the contributions of the interior and boundary residuals.
The parameters $\boldsymbol{\Theta}$ are then optimized via standard gradient-based algorithms such as Adam \cite{kingma2014adam} or L-BFGS \cite{liu1989limited} to minimize
$Loss_{total}$, thereby training the network to approximate the solution of Equation \eqref{eq:linear_PDEs}.


Although PINNs are applicable to a broad range of PDEs, their solutions often exhibit substantially lower accuracy than those obtained by classical numerical methods. Moreover, the dependence on gradient-based optimization of network parameters typically results in prolonged training times. Most critically, PINNs perform poorly on high-frequency problems, where the rapidly oscillatory solutions are inherently difficult for standard neural architectures to capture accurately.

\subsection{Randomized Neural Networks}
When solving linear PDEs, RNNs offer an alternative paradigm. In this framework, all network parameters except the output-layer weights are randomly initialized and kept fixed, yielding a predetermined set of basis functions. The PDE solution is then expressed as a linear combination of these basis functions, and the expansion coefficients are determined by solving a linear least-squares problem.

With all parameters randomly initialized and frozen, the neural network defines a fixed set of basis functions
\begin{equation}
    \boldsymbol{\Phi} = [\phi_1, \phi_2, ..., \phi_M]^T=\mathcal{N}(\boldsymbol{x}; \boldsymbol{\Theta}),
\end{equation}
where $M$ denotes the number of basis functions and $\boldsymbol{\Theta}$ collects the fixed network parameters. The approximate PDE solution is expressed as a linear combination of these basis functions,
\begin{equation}
    u_M = \boldsymbol{w}^T \boldsymbol{\Phi} = \sum_{i=1}^M w_i \phi_i,
\end{equation}
where the coefficient vector $\boldsymbol{w}=[w_1, w_2, ..., w_M]^T$ determined by solving a linear least-squares problem that enforces the PDE and boundary conditions at a set of collocation points.

Applying the linear differential operators 
$\mathcal{L}$ and $\mathcal{B}$ to $u_M$ yields
\begin{equation}
    \label{eq:u_M_operators}
    \begin{array}{r@{}l}
        \left\{
        \begin{aligned}
            \mathcal{L}u_M(\boldsymbol{x}) &= \sum_{i=1}^M w_i \mathcal{L}\phi_i(\boldsymbol{x}) = \boldsymbol{w}^T \mathcal{L}\boldsymbol{\Phi}, &  & \mbox{in} \enspace \Omega,          \\
            \mathcal{B}u_M(\boldsymbol{x}) &= \sum_{i=1}^M w_i \mathcal{B}\phi_i(\boldsymbol{x}) = \boldsymbol{w}^T \mathcal{B}\boldsymbol{\Phi},        &  & \mbox{on} \enspace \partial \Omega, \\
        \end{aligned}
        \right.
    \end{array}
\end{equation}
Assembling these expressions over the collocation sets $S_e$ and $S_b$ yields the block-structured matrix
\begin{equation}
    \label{eq:linear_system_left}
    \begin{aligned}
    \mathbf{A} &= 
    \begin{bmatrix}
        \mathcal{L}\boldsymbol{\Phi}(S_e) \\
        \mathcal{B}\boldsymbol{\Phi}(S_b) \\
    \end{bmatrix} \\
    &= 
    \begin{bmatrix}
        \mathcal{L}\phi_1(\boldsymbol{x}_1) & \mathcal{L}\phi_2(\boldsymbol{x}_1) & \cdots & \mathcal{L}\phi_M(\boldsymbol{x}_1) \\
        \vdots & \vdots & \vdots  & \vdots \\
        \mathcal{L}\phi_1(\boldsymbol{x}_{N_e}) & \mathcal{L}\phi_2(\boldsymbol{x}_{N_e}) & \cdots & \mathcal{L}\phi_M(\boldsymbol{x}_{N_e}) \\
        \mathcal{B}\phi_1(\boldsymbol{x}_{N_e+1}) & \mathcal{B}\phi_2(\boldsymbol{x}_{N_e+1}) & \cdots & \mathcal{B}\phi_M(\boldsymbol{x}_{N_e+1}) \\
        \vdots & \vdots & \vdots  & \vdots \\
        \mathcal{B}\phi_1(\boldsymbol{x}_{N_e+N_b}) & \mathcal{B}\phi_2(\boldsymbol{x}_{N_e+N_b}) & \cdots & \mathcal{B}\phi_M(\boldsymbol{x}_{N_e+N_b}) \\ 
    \end{bmatrix}.
    \end{aligned}
\end{equation}
As described in the previous section, the coefficient vector $\boldsymbol{w}$ is then obtained by solving the linear least-squares problem $\mathbf{A} \boldsymbol{w} = \mathbf{F}$, where the right-hand side vector $\mathbf{F}$ is defined as
\begin{equation}
    \label{eq:linear_system_right}
    \boldsymbol{F} = 
    \begin{bmatrix}
    f(\boldsymbol{x}_1) \\
    \vdots \\
    f(\boldsymbol{x}_{N_e}) \\
    g(\boldsymbol{x}_{N_e+1}) \\
    \vdots \\
    g(\boldsymbol{x}_{N_e + N_b +1}) \\ 
    \end{bmatrix},
\end{equation}
where $f(\boldsymbol{x}_i)$ for $ i = 1, ...,  N_e$ denotes the source term evaluated at the ‌interior collocation points, and $g(\boldsymbol{x}_i)$ for $i = N_e+1, ...,  N_e+N_b$ are the prescribed boundary values at the boundary collocation points.

RNNs have attracted considerable interest for solving PDEs, as they deliver accurate solutions without iterative training at high computational efficiency. Among common architectures---fully-connected, residual, and single-hidden-layer neural networks (SHLNNs)---SHLNNs are the predominant choice because of their simplicity and closed-form derivatives of basis functions, which eliminate the need for AD. Nevertheless, RNNs still struggle to resolve solutions with high-frequency content or sharp spatial oscillations.

\section{Multi-Scale Separable Fourier Neural Networks}
\label{sec:methods}
 Both PINNs and RNNs struggle to resolve high-frequency PDE solutions. To address this limitation, we propose the Multi-Scale Separable Fourier Neural Network (MS-SFNN), which integrates three key components: variable separation, per-dimension adaptive frequency scaling, and Fourier feature embedding via cosine activation.

We adopt a variable-separation strategy. For an input $\boldsymbol{x} \in \mathbb{R}^d$,  we employ $d$ independent SHLNNs, each processing a single coordinate. The  $j$-th subnetwork is 
\begin{equation}
    \mathcal{N}_j(x_j; \boldsymbol{\Theta}_j) = \sigma(\boldsymbol{w}_j x_j + \boldsymbol{b}_j), \enspace j=1, 2, \cdots, d,
\end{equation}
where $\boldsymbol{w}_j \in \mathbb{R}^{M}$ and $\boldsymbol{b}_j \in \mathbb{R}^M$ fixed, randomly initialized parameters, each entry drawn i.i.d. from a zero-mean unit-variance uniform distribution. The $j$-th subnetwork outputs the vector $\boldsymbol{\phi}_j = \mathcal{N}_j(x_j, \boldsymbol{\Theta}_j) = [\phi_{j1}, \phi_{j2}, \cdots, \phi_{jM}]^T$. The global basis functions are formed by the element-wise product of all $d$ subnet outputs:
\begin{equation}
    \begin{aligned}
    \boldsymbol{\Phi} &= \prod_{j=1}^d \boldsymbol{\phi}_j = \prod_{j=1}^d \mathcal{N}_j(x_j; \boldsymbol{\Theta}_j) = \prod_{j=1}^d \sigma(\boldsymbol{w}_j x_j + \boldsymbol{b}_j).
    \end{aligned}
\end{equation}
Consequently, the approximate PDE solution is 
\begin{equation}
    \label{eq:SV_uM}
    \begin{aligned}
    u_M(\boldsymbol{x}) &= \boldsymbol{w}^T \boldsymbol{\Phi} = \sum_{i=1}^M w_i \prod_{j=1}^d \sigma(w_{ji} x_j + b_{ji}). 
    \end{aligned}
\end{equation}

\begin{figure}[htbp]
    \centering
    \begin{minipage}{1.0\linewidth}
        \includegraphics[width=1\textwidth]{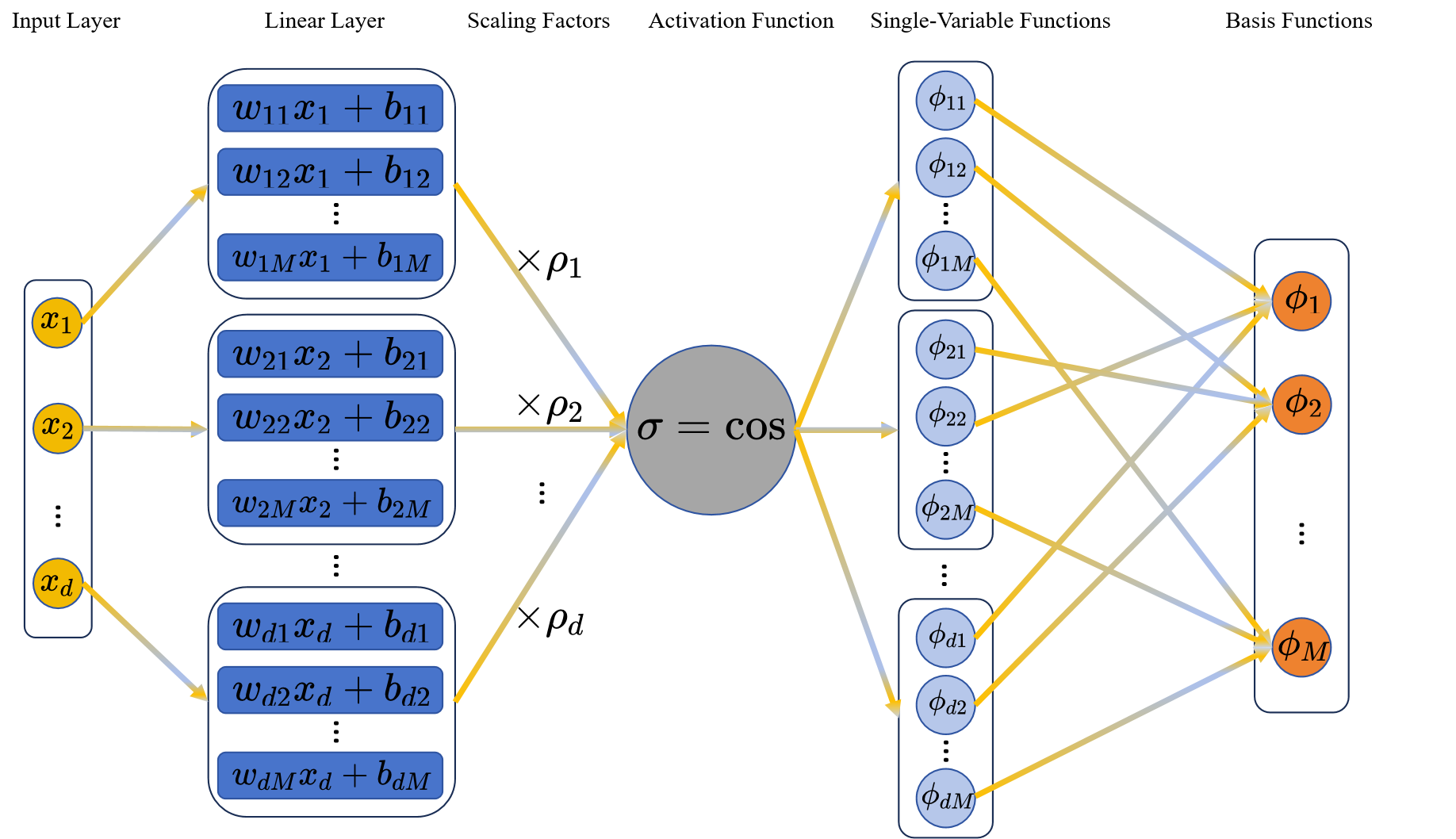}
    \end{minipage}    
    \caption{Multi-Scale Separable Fourier Neural Networks (MS-SFNN) architecture diagram. This architecture employs variable separation: each input dimension $x_j$ is processed independently by a dedicated single-hidden-layer network. Each subnetwork applies a linear map, followed by a dimension-specific scaling factor $\rho_j$ and a cosine activation. The $M$-dimensional outputs from all subnetworks are combined via element-wise multiplication to form the basis functions. The final solution is a linear combination of these bases.}
    \label{fig:MS-SFNN}
\end{figure}

Although the variable-separation representation in Equation \eqref{eq:SV_uM} decouples the input dimensions, fixed unit-variance weights and biases alone lack the spectral capacity to resolve high-frequency features. Prior work \cite{FENs} demonstrated that parameter scaling can enhance representational power, yet a uniform global factor $\rho$ proves insufficient for high-frequency problems, because distinct input variables typically demand different frequency emphasis.
We therefore assign a tunable per-dimension scaling factor $\rho_j$ to each subnetwork, modifying Eq.~\eqref{eq:SV_uM} to
\begin{equation}
    \label{eq:SV_uM_rho}
    u_M(\boldsymbol{x}) = \sum_{i=1}^M w_i \prod_{j=1}^d \sigma\left(\rho_j(w_{ji} x_j + b_{ji})\right),
\end{equation}
where $\rho_j$ modulates the frequency content along the 
$j$-th coordinate. Figure \ref{fig:MS-SFNN} illustrates the complete MS-SFNN architecture.

Additionally, we forgo conventional sigmoidal activations (sigmoid, tanh, swish) in favor of the cosine function. As shown in prior work \cite{FENs}, trigonometric activations intrinsically embed Fourier features and offer distinctly better approximation for oscillatory solutions. Since 
$\cos$, $\sin$, and their combinations deliver comparable accuracy for a given number of basis functions, we adopt $\cos$ for simplicity. The approximate solution then becomes
\begin{equation}
    \label{eq:SV_uM_rho_cos}
    u_M(\boldsymbol{x}) = \sum_{i=1}^M w_i \prod_{j=1}^d \cos(\rho_j(w_{ji} x_j + b_{ji})).
\end{equation}


A key advantage of MS-SFNN is that it entirely avoids AD. 
The explicit separable form of  $u_M$ in \eqref{eq:SV_uM_rho_cos} yields closed-form spatial derivatives. For instance, the first- and second-order partial derivatives with respect to $x_k$ are
\begin{align}
    & \frac{\partial u_M}{\partial x_k}  =  \sum_{i=1}^M w_i \prod_{j \neq k}^d \cos\left(\rho_j(w_{ji} x_j + b_{ji})\right)(-\rho_k w_{ki})\sin\left(\rho_k(w_{ki} x_k + b_{ki})\right), \label{eq:u_xk} \\
   &  \frac{\partial^2 u_M}{\partial x_k^2}  = \sum_{i=1}^M w_i \prod_{j=1}^d \cos(\rho_j(w_{ji} x_j + b_{ji}))(-\rho_k^2 w_{ki}^2).  \label{eq:u_xkxk}
\end{align}
These analytical expressions enable exact derivative computation with minimal overhead, completely eliminating the memory burden and numerical noise inherent in AD.

\section{Numerical Experiments}
\label{sec:experiments}
In this section, we present numerical experiments to demonstrate the applicability and accuracy of the proposed MS-SFNN. 
GPU memory constitutes the principal bottleneck in high-frequency settings, stemming from assembling the system matrix 
$\mathbf{A}$ (Equation \eqref{eq:linear_system_left}) and subsequently solving the least-squares system for $\boldsymbol{w}$. Since the solution is fully determined by $\mathbf{A}$, we focus on its construction. Substituting AD with the analytical derivatives in \eqref{eq:u_xk}--\eqref{eq:u_xkxk} eliminates the computational graph, yet when 
$N$ reaches the order of millions even the analytical evaluation of $\mathcal{L}\boldsymbol{\Phi}$ and  $\mathcal{B}\boldsymbol{\Phi}$ exceeds GPU memory. We address this using batched QR decomposition \cite{chen_pyrfm_2026}, which incrementally processes small collocation batches to build only an $M \times M$ upper-triangular matrix, dramatically reducing peak memory while preserving full accuracy.

We conducted all experiments on a Debian 12 server equipped with an Intel Xeon Platinum 8358 CPU (2.60 GHz) and an NVIDIA A100 GPU (80 GB). 
To quantitatively evaluate the approximation capabilities of neural networks in the numerical experiments, the maximum absolute error ($L_{\infty}$ error) and the relative $L_2$ error are defined as follows:
\begin{eqnarray}
    \label{eq:definitionerror}
    e_{L_{\infty}}  = \max_{1\leq i \leq N} \vert u_M(\boldsymbol{x}_i)-u_{exact}(\boldsymbol{x}_i) \vert,  \\
    e_{L_2} = \sqrt{\frac{\sum_{i=1}^{N}(u_M(\boldsymbol{x}_i)-u_{exact}(\boldsymbol{x}_i))^2}{\sum_{i=1}^{N}(u_{exact}(\boldsymbol{x}_i))^2}},
\end{eqnarray}
where $u_M$ and $u_{\text{exact}}$ denote the approximate and exact solutions, respectively, and $\boldsymbol{x}_i$ ($1 \le i \le N$) are the collocation points used for error evaluation.
Unless stated otherwise, the number of basis functions is fixed at $M=10{,}000$.

\subsection{Two-Dimensional Helmholtz Equations}
\label{sec4.1}
We evaluate MS-SFNN on the two-dimensional Helmholtz equation with high-frequency oscillations:
\begin{align}
            \Delta u + k^2 u  & =   f,  \  \mbox{in} \enspace \Omega,  \label{eq:Helmholtz_Equation}\\
             u             & =  0,   \  \mbox{on} \enspace \partial \Omega,  \label{eq:Helmholtz_EquationDich}
\end{align}
where $\Omega = (0,1)\times(0,1)$, $u=\sin(k x)\sin(k y)$, $f= -k^2 \sin(k x)\sin(k y)$ and $k=24\pi$, $48\pi$.

\begin{figure}[htbp]
    \centering
    \begin{minipage}{1.0\linewidth}
        \includegraphics[width=1\textwidth]{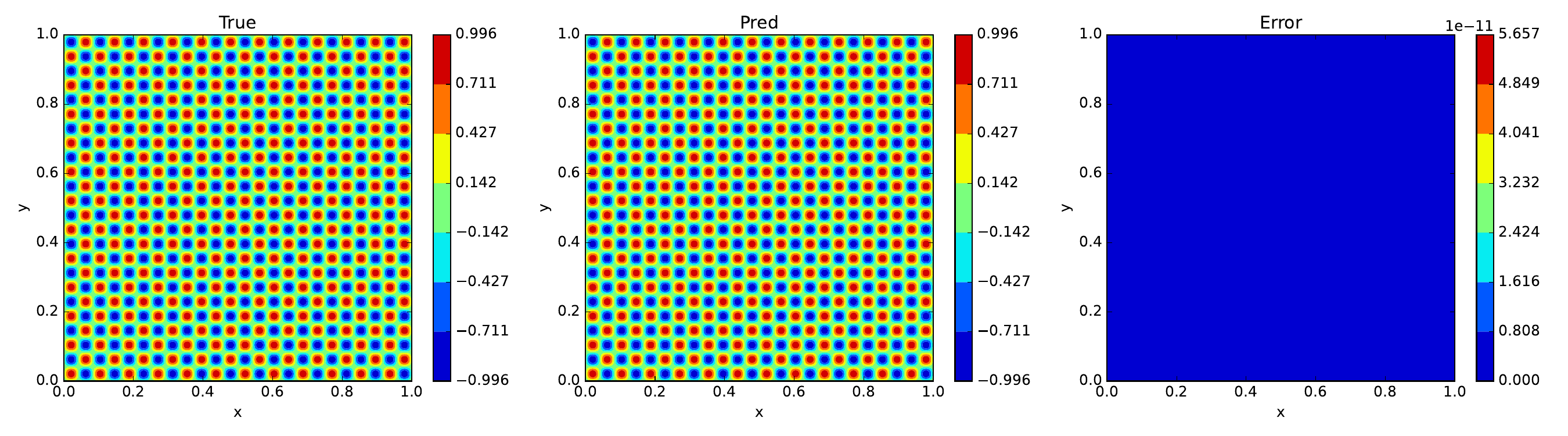}
    \end{minipage}
    \caption{Heat maps illustrate the two-dimensional Helmholtz equation \eqref{eq:Helmholtz_Equation} with $k=24 \pi$. Left: the exact solution; Middle: the prediction solution of MS-SFNN; Right: the absolute error between them.}
    \label{fig:HelmholtzEquations_24pi}
\end{figure}

\begin{figure}[htbp]
    \centering
    \begin{minipage}{1.0\linewidth}
        \includegraphics[width=1\textwidth]{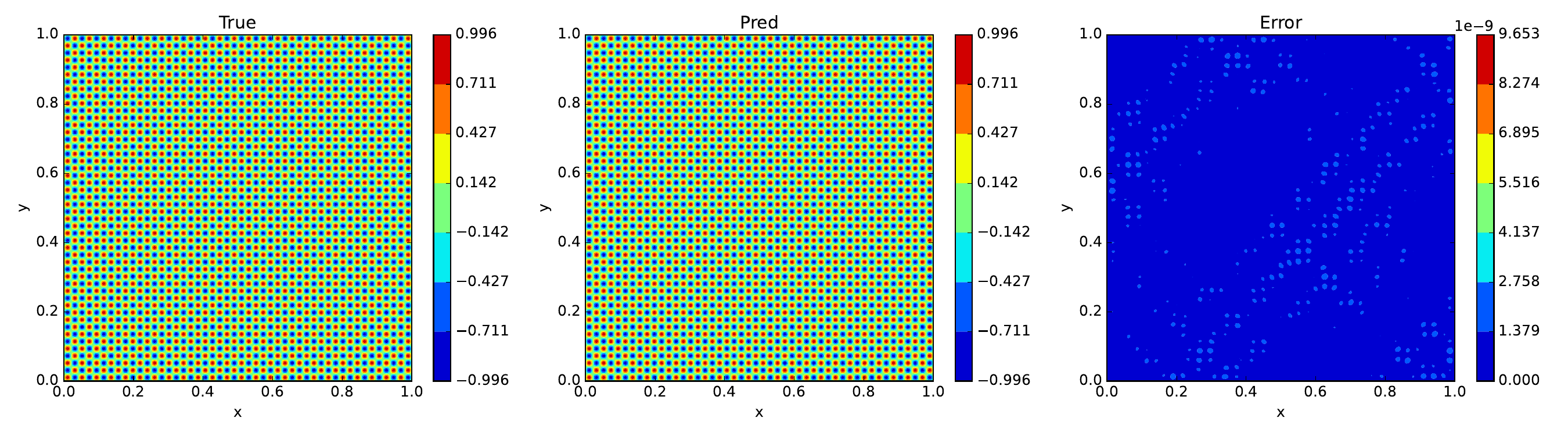}
    \end{minipage}
    \caption{Heat maps illustrate the two-dimensional Helmholtz equation \eqref{eq:Helmholtz_Equation} with $k=48 \pi$. Left: the exact solution; Middle: the prediction solution of MS-SFNN; Right: the absolute error between them.}
    \label{fig:HelmholtzEquations_48pi}
\end{figure}

The equation is solved on a uniform $N_x \times N_y = 201 \times 201$ grid of training points. 
For the 2D case, MS-SFNN employs two scaling factors, set to $\rho_1=\rho_2=24\pi$ for $k=24\pi$ and $\rho_1=\rho_2=48\pi$ for $k=48\pi$.
Table \ref{tab:2D_Helmholtz_Equations} lists the  $L_{\infty}$ and $L_2$ errors for PINN, SV-SNN, and MS-SFNN. It can be observed that PINN fails to solve the problem at both wave numbers, while SV-SNN yields errors around $10^{-2}$. In sharp contrast, MS-SFNN achieves high accuracy; for $k = 48 \pi$, the $L_{\infty}$ and $L_2$ errors reach $9.65 \times 10^{-9}$ and $1.34 \times 10^{-9}$, respectively. Figures \ref{fig:HelmholtzEquations_24pi} and \ref{fig:HelmholtzEquations_48pi} display heat maps of the exact solution, the MS-SFNN prediction, and the absolute error, confirming the method’s effectiveness with negligible error levels.

\begin{table}[htp]
    \begin{center}
        \caption{Two-dimensional Helmholtz equations \eqref{eq:Helmholtz_Equation}--\eqref{eq:Helmholtz_EquationDich}: Performance comparison of PINN, SV-SNN and MS-SFNN. The $L_{\infty}$ errors and $L_{2}$ errors for each model configuration are presented.}
        \begin{tabular}{ccccccc}
            \hline\noalign{\smallskip}
            \multirow{2}{*}{Method} & \multicolumn{2}{c}{$k=24\pi$} & \multicolumn{2}{c}{$k=48\pi$} \\
            & $e_{L_{\infty}}$ & $e_{L_{2}}$ & $e_{L_{\infty}}$ & $e_{L_{2}}$ \\
            \hline
            \text{PINN}\cite{PINN}    & 1.09E+00 & 1.01E+00 & 1.09E+00 & 1.00E+00 \\
            \text{SV-SNN}\cite{SV-SNN}   & 3.62E-02 & 1.33E-02 & 5.49E-02 & 3.99E-03 \\
            \text{MS-SFNN}  & 5.66E-11 & 2.71E-12 & 9.65E-09 & 1.34E-09 \\
            \hline
        \end{tabular}
        \label{tab:2D_Helmholtz_Equations}
    \end{center}
\end{table}

\subsection{Complex Geometry Helmholtz Equations}
We now assess MS-SFNN on geometrically complex domains using the 2D Helmholtz equation in a square with a cylindrical obstacle. The domain is 
$\Omega=(0,1)\times(0,1) \backslash \Omega_c$, where $\Omega_c$ is a cylinder of radius 
$r = 0.15$ centered at $(0.5,0.5)$. The governing equation is the same as in Equation \eqref{eq:Helmholtz_Equation}, with Dirichlet conditions $u = 0$ on the outer square boundary and  $u=\sin(kx)\sin(ky)$ on the cylinder surface.  Training points are obtained from a uniform 
$201 \times 201$ grid by discarding points inside or on the obstacle; the retained grid provides interior collocation and outer boundary points, while the inner boundary is discretized with 
$200$ uniformly spaced points in polar coordinates. The scaling factors are set to 
$\rho_1=\rho_2=24\pi$ for $k = 24 \pi$, and $\rho_1=\rho_2=48\pi$ for $k = 48 \pi$.

\begin{figure}[htbp]
    \centering
    \begin{minipage}{1.0\linewidth}
        \includegraphics[width=1\textwidth]{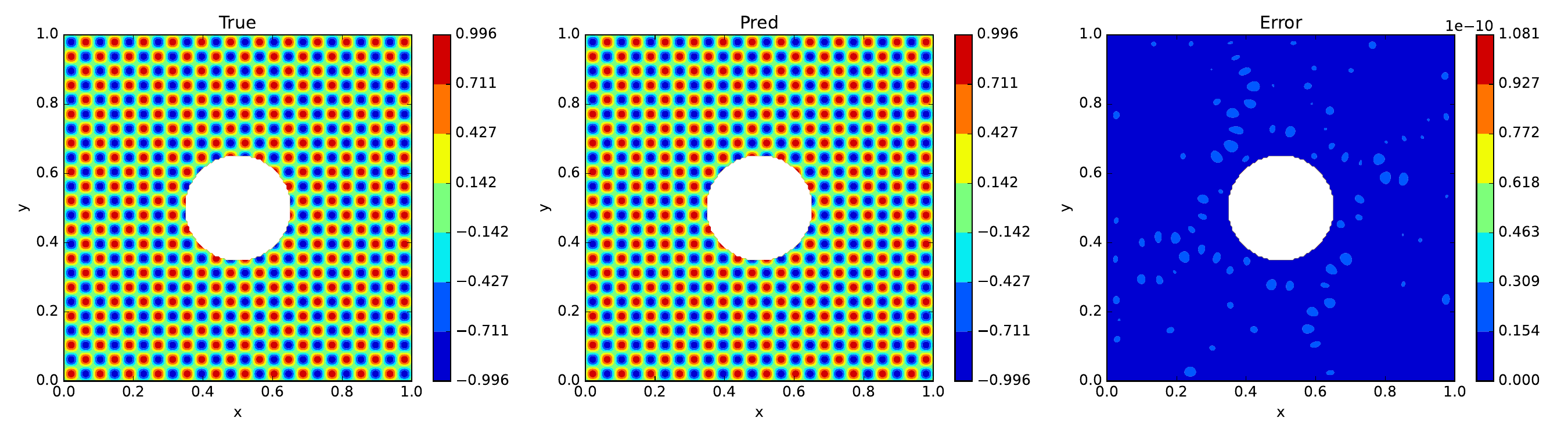}
    \end{minipage}
    \caption{Heat maps illustrate the two-dimensional complex geometry Helmholtz equation with $k=24 \pi$. Left: the exact solution; Middle: the prediction solution of MS-SFNN; Right: the absolute error between them.}
    \label{fig:CG_HelmholtzEquations_24pi}
\end{figure}

\begin{figure}[htbp]
    \centering
    \begin{minipage}{1.0\linewidth}
        \includegraphics[width=1\textwidth]{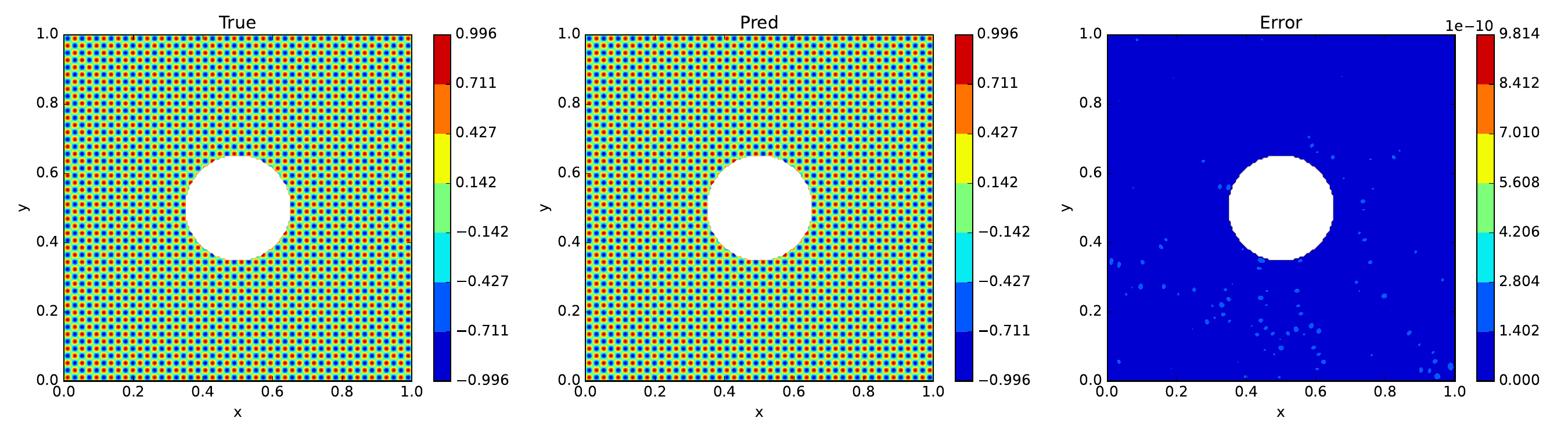}
    \end{minipage}
    \caption{Heat maps illustrate the two-dimensional complex geometry Helmholtz equation with $k=48 \pi$. Left: the exact solution; Middle: the prediction solution of MS-SFNN; Right: the absolute error between them.}
    \label{fig:CG_HelmholtzEquations_48pi}
\end{figure}

As shown in Table \ref{tab:CG_Helmholtz_Equation}, for $k=24\pi$, PINN fails entirely, while SV-SNN reaches only $L_{\infty}$ error of $1.24 \times 10^{-1}$, highlighting the severe difficulty high-frequency problems pose even for specialized architectures.
In marked contrast,  the proposed MS-SFNN solves the Helmholtz equation on this complex geometry with exceptional accuracy, reaching error levels on the order of $10^{-10}$. Figures \ref{fig:CG_HelmholtzEquations_24pi} and \ref{fig:CG_HelmholtzEquations_48pi} display the exact solution, the MS-SFNN approximation, and the corresponding absolute error for $k=24\pi$ and $k=48\pi$, respectively, confirming the effectiveness of MS-SFNN.

\begin{table}[htp]
    \begin{center}
        \caption{Complex geometry Helmholtz equations: Performance comparison of SV-SNN and MS-SFNN. The $L_{\infty}$ errors and $L_{2}$ errors for each model configuration are presented.}
        \begin{tabular}{ccccccc}
            \hline\noalign{\smallskip}
            \multirow{2}{*}{Method} & \multicolumn{2}{c}{$k=24\pi$} & \multicolumn{2}{c}{$k=48\pi$} \\
            & $e_{L_{\infty}}$ & $e_{L_{2}}$ & $e_{L_{\infty}}$ & $e_{L_{2}}$ \\
            \hline
            \text{PINN}\cite{PINN}    & 1.64E+00 & - & - & - \\
            \text{SV-SNN}\cite{SV-SNN}   & 1.24E-01 & - & - & - \\
            \text{MS-SFNN}  & 1.08E-10 &1.49E-11 & 9.81E-10 & 1.08E-10 \\
            \hline
        \end{tabular}
        \label{tab:CG_Helmholtz_Equation}
    \end{center}
\end{table}

\subsection{Complex Geometry Poisson Equations}
We further evaluate MS-SFNN on the Poisson equation in a multiply perforated domain, governed by
\begin{eqnarray}
            -\Delta u  & = f, &      \mbox{in} \enspace \Omega,  \label{eq:CG_Poisson_Equation} \\
            u             & = g,       & \mbox{on} \enspace \partial \Omega,  \label{eq:CG_Poisson_EquationDich}
\end{eqnarray}
where $\Omega=(-1,1)^2$, $u=\sin(\mu x) \sin(\mu y)$, $f=2\mu^2 \sin(\mu x)\sin(\mu y)$, $g=u(x, y)$, $\mu = 7\pi$. The complex domain configuration includes external boundary and multiple internal holes: domain contains three circular holes located at $(-0.5, -0.5)$, $(0.5, 0.5)$, $(0.5, -0.5)$ with radius $0.1$, $0.2$, $0.2$, respectively and one ellipse with equation $16(x+0.5)^2+64(y-0.5)^2=1$.

We construct training data from a uniform 
$201 \times 201$ grid, discarding points inside or on the hole boundaries to yield interior and outer boundary points. Each hole boundary is discretized with $200$ uniformly spaced points in polar coordinates. For this 2D Poisson problem, two scaling factors are set to  $21$. 

\begin{figure}[htbp]
    \centering
    \begin{minipage}{1.0\linewidth}
        \includegraphics[width=1\textwidth]{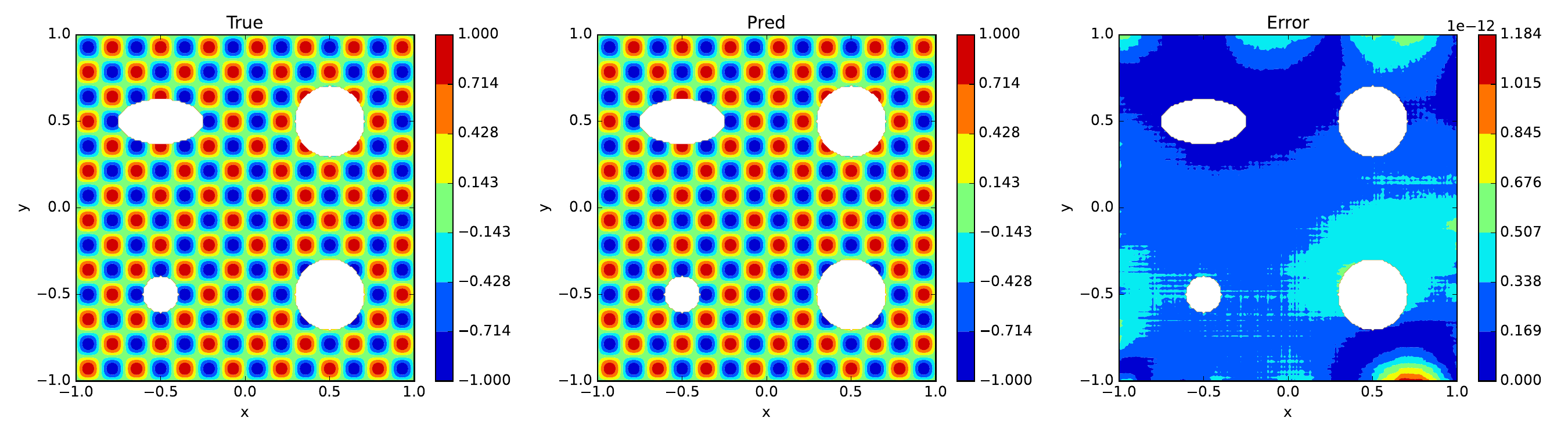}
    \end{minipage}
    \caption{Heat maps illustrate the complex geometry Poisson equation \eqref{eq:CG_Poisson_Equation}. Left: the exact solution; Middle: the prediction solution of MS-SFNN; Right: the absolute error between them.}
    \label{fig:ComplexGeometryPoissonEquations}
\end{figure}

As Table~\ref{tab:CG_Poisson_Equation} shows, SV-SNN achieves an  $L_{\infty}$  error of $3.45 \times 10^{-2}$, whereas MS-SFNN reaches $1.18 \times 10^{-12}$, an improvement of ten orders of magnitude. Figure \ref{fig:ComplexGeometryPoissonEquations} displays the exact solution, the MS-SFNN approximation, and the absolute error, confirming that MS-SFNN attains high accuracy even in a domain with four internal holes.

\begin{table}[htp]
    \begin{center}
        \caption{Complex geometry Poisson equations \eqref{eq:CG_Poisson_Equation}--\eqref{eq:CG_Poisson_EquationDich}: Performance comparison of SV-SNN and MS-SFNN. The $L_{\infty}$ errors and $L_{2}$ errors for each model configuration are presented.}
        \begin{tabular}{ccccccc}
            \hline\noalign{\smallskip}
            Method & $e_{L_{\infty}}$ & $e_{L_{2}}$ \\
            \hline
            \text{SV-SNN}\cite{SV-SNN}   & 3.45E-02 & - \\
            \text{MS-SFNN}  & 1.18E-12 & 5.75E-13 \\
            \hline
        \end{tabular}
        \label{tab:CG_Poisson_Equation}
    \end{center}
\end{table}

\subsection{Two-Dimensional Flower-Shaped Problems with Mixed Boundary}
Further, we consider a two-dimensional flower-shaped problem with complex boundary conditions. The problem is given by \eqref{eq:Helmholtz_Equation} with following boundary conditions
\begin{equation}
    \label{eq:Flower_Mixed_Equation}
    \begin{array}{r@{}l}
        \left\{
        \begin{aligned}
            u             & = g_1,        &  & \mbox{on} \enspace \Gamma_1, \\
            u_n             & = g_2,        &  & \mbox{on} \enspace \Gamma_2, \\
            u_n+u             & = g_3,        &  & \mbox{on} \enspace \Gamma_3, \\
        \end{aligned}
        \right.
    \end{array}
\end{equation}
where the boundary is parameterized as $r = a - b\cos(m\theta)$ and the sub-boundaries $\Gamma_1$, $\Gamma_2$ and $\Gamma_3$ correspond to $\theta \in [0,\frac{2}{3}\pi)$, $\theta \in [\frac{2}{3}\pi,\frac{4}{3}\pi)$, and $\theta \in [\frac{4}{3}\pi, 2\pi)$, respectively.
The exact solution $u_{exact}=\sin(\frac{k}{\sqrt{2}} x)\sin(\frac{k}{\sqrt{2}} y)$, with the wavenumber $k=\frac{2\pi \nu_{freq}}{340}$, where the frequency takes values $\nu_{freq}=1{,}000$ and $10{,}000$. The corresponding source term $f$ and right term $g_1$, $g_2$ and $g_3$ are derived from the exact solution.

Training points are generated from a uniform $201 \times 201$ grid; points outside or on the flower-shaped boundary are discarded, retaining only interior points. For the boundary, $40{,}000$ points are uniformly sampled on the curve and partitioned into $\Gamma_1$, $\Gamma_2$, $\Gamma_3$ according to the angle intervals above. 
MS-SFNN uses scaling factors $\rho_1=\rho_2=\frac{k}{\sqrt{2}}$.

\begin{figure}[htbp]
    \centering
    \begin{minipage}{1.0\linewidth}
        \includegraphics[width=1\textwidth]{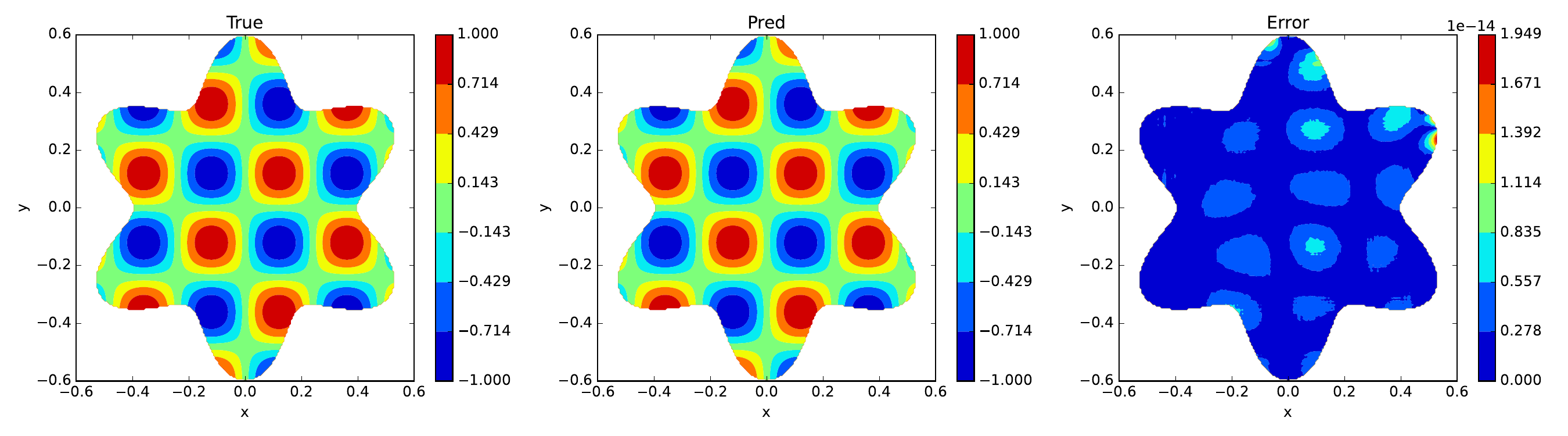}
    \end{minipage}
    \caption{Heat maps illustrate the two-dimensional flower-shaped problems with mixed boundary \eqref{eq:Flower_Mixed_Equation} with $\nu_{freq}=1{,}000$Hz. Left: the exact solution; Middle: the prediction solution of MS-SFNN; Right: the absolute error between them.}
    \label{fig:FlowerShape_2D_Mixed_Boundary_1000}
\end{figure}
\begin{figure}[htbp]
    \centering
    \begin{minipage}{1.0\linewidth}
        \includegraphics[width=1\textwidth]{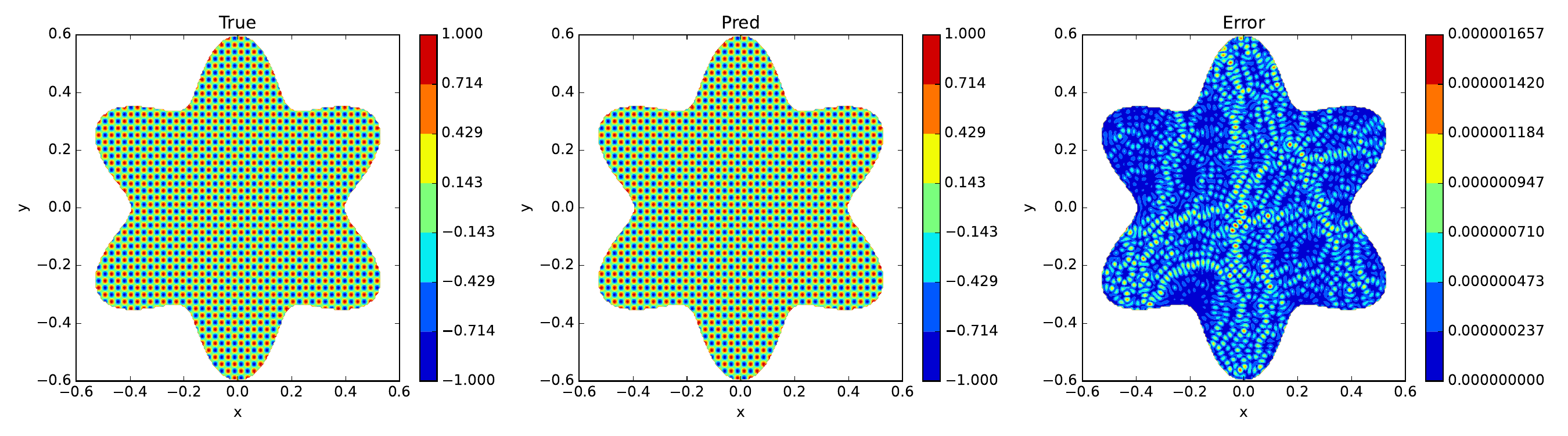}
    \end{minipage}
    \caption{Heat maps illustrate the two-dimensional flower-shaped problems with mixed boundary \eqref{eq:Flower_Mixed_Equation} with $\nu_{freq}=10{,}000$Hz. Left: the exact solution; Middle: the prediction solution of MS-SFNN; Right: the absolute error between them.}
    \label{fig:FlowerShape_2D_Mixed_Boundary_10000}
\end{figure}

For frequency $v_{freq}=1{,}000$, FEM achieves $e_{L_{\infty}} = 2.60 \times 10^{-3}$, LbNM reaches $2.57 \times 10^{-6}$, whereas MS-SFNN attains $e_{L_{\infty}} = 1.95 \times 10^{-14} $ and $e_{L_{2}} = 5.33 \times 10^{-15}$. At $v_{feq}=10{,}000$, FEM fails, LbNM gives $e_{L_{\infty}} = 2.66 \times 10^{-6}$, and MS-SFNN again delivers significantly higher accuracy. The comparisons are shown in Table \ref{tab:Flower_Mixed_Equation}. Figures~\ref{fig:FlowerShape_2D_Mixed_Boundary_1000} and~\ref{fig:FlowerShape_2D_Mixed_Boundary_10000} show the exact solution, the MS-SFNN approximation, and the absolute error for both frequencies, confirming the excellent performance.

\begin{table}[htp]
    \begin{center}
        \caption{Two-dimensional flower-shaped problems with mixed boundary \eqref{eq:Flower_Mixed_Equation}: Performance comparison of LbNM and MS-SFNN. The $L_{\infty}$ errors and $L_{2}$ errors for each model configuration are presented.}
        \begin{tabular}{ccccccc}
            \hline\noalign{\smallskip}
            \multirow{2}{*}{Method} & \multicolumn{2}{c}{$\nu_{freq}=1{,}000$Hz} & \multicolumn{2}{c}{$\nu_{freq}=10{,}000$Hz} \\
            & $e_{L_{\infty}}$ & $e_{L_{2}}$ & $e_{L_{\infty}}$ & $e_{L_{2}}$ \\
            \hline
            \text{LbNM}\cite{LI2024200}  & 2.57E-06 & - & 2.66E-06 & - \\
            \text{MS-SFNN}  & 1.95E-14 & 5.33E-15 & 1.66E-06 & 8.37E-07 \\
            \hline
        \end{tabular}
        \label{tab:Flower_Mixed_Equation}
    \end{center}
\end{table}

\subsection{Three-Dimensional Helmholtz Equations}
In this example, we consider the three-dimensional Helmholtz equation \eqref{eq:Helmholtz_Equation} in the domain $\Omega=\left\{(x,y,z) \vert x^2+y^2+z^2 \leq 1 \right\}$ with boundary condition \eqref{eq:Helmholtz_EquationDich}, 
where the exact solution is given by $$u_{exact}=\sin(x^2+y^2+z^2-1) \cos(6x)\cos(ky)(z^2+1),$$ and $k=10.5\pi$, the corresponding forcing term $f$ can be obtained using the exact solution.

For the three-dimensional test, we employ a uniform  $101 \times 101 \times 101$ grid, etaining interior points within the sphere for the PDE and sampling the spherical boundary with $500 \times 500$ uniformly spaced points in spherical coordinates. 
MS-SFNN uses scaling factors $\rho_1=6$, $\rho_2=10.5\pi$, $\rho_3=2$.

\begin{figure}[htbp]
    \centering
    \begin{minipage}{1.0\linewidth}
        \includegraphics[width=1\textwidth]{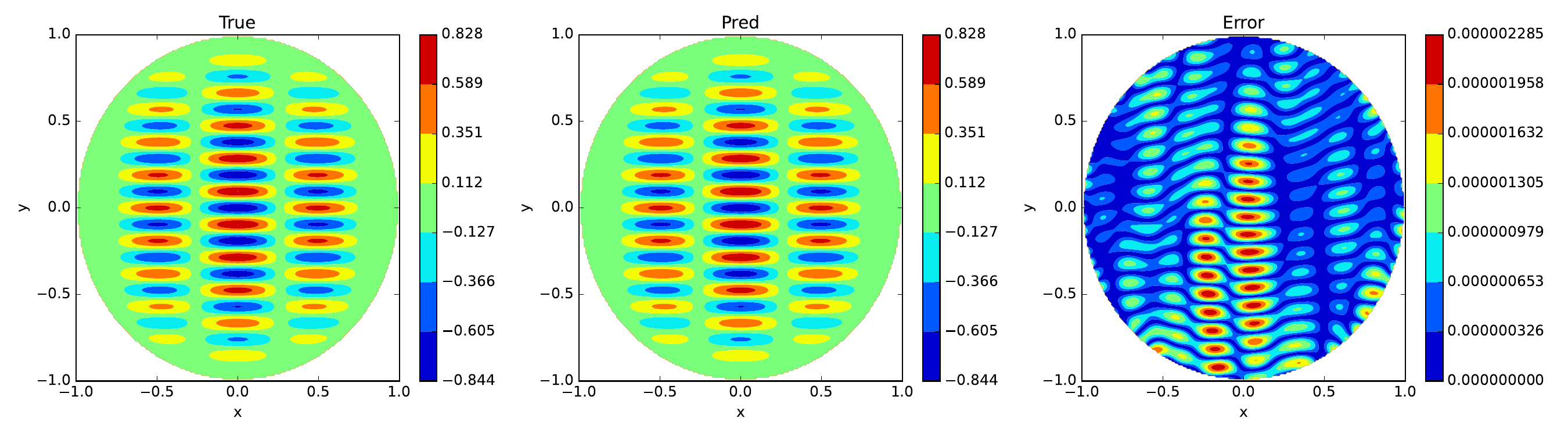}
    \end{minipage}
    \caption{Heat maps illustrate the three-dimensional Helmholtz equation \eqref{eq:Helmholtz_Equation}--\eqref{eq:Helmholtz_EquationDich} at $z=0.1$. Left: the exact solution; Middle: the prediction solution of MS-SFNN; Right: the absolute error between them.}
    \label{fig:3D_Helmholtz_Equations}
\end{figure}

\begin{figure}[htbp]
    \centering
    \begin{minipage}{1.0 \linewidth}
        \includegraphics[width=1\textwidth]{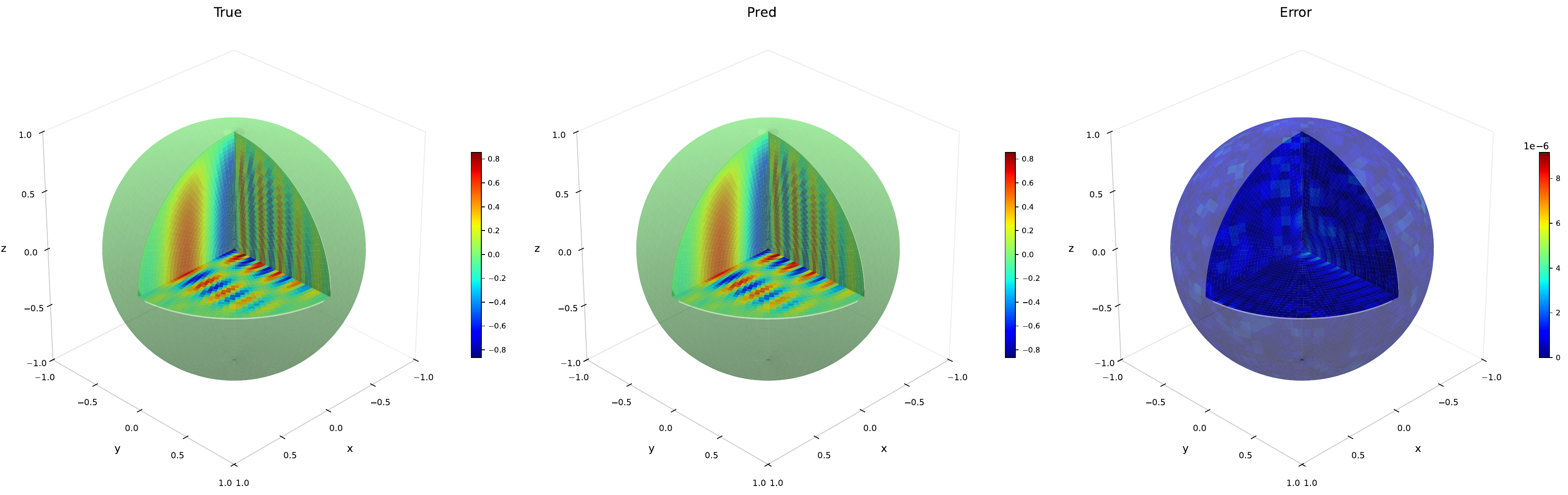}
    \end{minipage}
    \caption{Heat maps illustrate the three-dimensional Helmholtz equation \eqref{eq:Helmholtz_Equation}--\eqref{eq:Helmholtz_EquationDich}. Left: the exact solution; Middle: the prediction solution of MS-SFNN; Right: the absolute error between them.}
    \label{fig:3D_Helmholtz_Equations_sphere}
\end{figure}

We compare MS-SFNN against PINN, SR-HFNN, and NR-HFNN. As Table \ref{tab:3D_Helmholtz_Equations} shows, PINN fails, while NR-HFNN and SR-HFNN yield
$L_2$ errors of $7.30 \times 10^{-3}$ and $9.04 \times 10^{-4}$, respectively. In contrast, MS-SFNN achieves $e_{L_{\infty}} = 3.30 \times 10^{-6}$ and $e_{L_{2}} = 2.94 \times 10^{-6}$, substantially outperforming the other methods. Figure \ref{fig:3D_Helmholtz_Equations} displays a cross-section at $z = 0.1$, and Figure \ref{fig:3D_Helmholtz_Equations_sphere} visualizes the exact solution, MS-SFNN prediction, and absolute error in 3D with cutaway slices, confirming the method's high accuracy on this spherical-domain problem.

\begin{table}[htp]
    \begin{center}
        \caption{Three-dimensional Helmholtz equation \eqref{eq:Helmholtz_Equation}--\eqref{eq:Helmholtz_EquationDich}: Performance comparison of PINN, SR-HFNN, NR-HFNN and MS-SFNN. The $L_{\infty}$ errors and $L_{2}$ errors for each model configuration are presented.}
        \begin{tabular}{ccccccc}
            \hline\noalign{\smallskip}
            Method & $e_{L_{\infty}}$ & $e_{L_{2}}$ \\
            \hline
            \text{PINN}\cite{PINN}    & - & 2.87E+00 \\
            \text{SR-HFNN}\cite{SR-HFNN}   & - & 9.04E-04 \\
            \text{NR-HFNN}\cite{SR-HFNN}   & - & 7.30E-03 \\
            \text{MS-SFNN}  & 3.30E-06 & 2.94E-06 \\
            \hline
        \end{tabular}
        \label{tab:3D_Helmholtz_Equations}
    \end{center}
\end{table}

\subsection{Nonlinear Elliptic Equations}
To evaluate MS-SFNN on nonlinear PDEs, we consider the two-dimensional nonlinear elliptic equation
\begin{equation}
    \label{eq:NonlinearEllipticEquations}
    \begin{array}{r@{}l}
        \left\{
        \begin{aligned}
            \Delta u + u^2  & = f, &  &  \mbox{in} \enspace\Omega, \\
            u             & = g,        &  & \mbox{on} \enspace \partial \Omega, 
        \end{aligned}
        \right.
    \end{array}
\end{equation}
with $\Omega=(0,1) \times (0,1)$ and exact solution  $u=(x+y)\cos(k x)\sin(k y)$, $k=10$, from which $f$ and $g$ are derived. Training points are taken from a uniform $101 \times 101$ grid serving as collocation points for both the PDE and the boundary condition. 
We use scaling factors $\rho_1=\rho_2=10$. Because the PDE is nonlinear, the resulting algebraic system is solved by Picard iteration.

The Picard iteration for \eqref{eq:NonlinearEllipticEquations} proceeds as follows. An initial coefficient vector $\boldsymbol{w}_0$  is generated via Xavier initialization, yielding $u_0(x, y) = \boldsymbol{\Phi}(x, y) \cdot \boldsymbol{w}_0$. At iteration
$i \ge 1$,  the nonlinear term is frozen at the previous iterate $u_{i-1}$, leading to the linearized equation $\Delta u+ u_{i-1} u = f$. This linear boundary-value problem is solved using the MS-SFNN least-squares framework, producing a new coefficient vector $\boldsymbol{w}_{i}$ and and approximation $u_{i}(x, y)= \boldsymbol{\Phi}(x, y) \cdot \boldsymbol{w}_{i}$.  The process repeats until $\lVert \boldsymbol{w}_{i} - \boldsymbol{w}_{i-1} \rVert \le 10^{-16}$ or a maximum of $100$ iterations is reached. The converged coefficients $\boldsymbol{w}^*$ define the high-accuracy solution $u^*(x, y) = \boldsymbol{\Phi}(x, y) \cdot \boldsymbol{w}^*$.

\begin{figure}[htbp]
    \centering
    \begin{minipage}{1.0\linewidth}
        \includegraphics[width=1\textwidth]{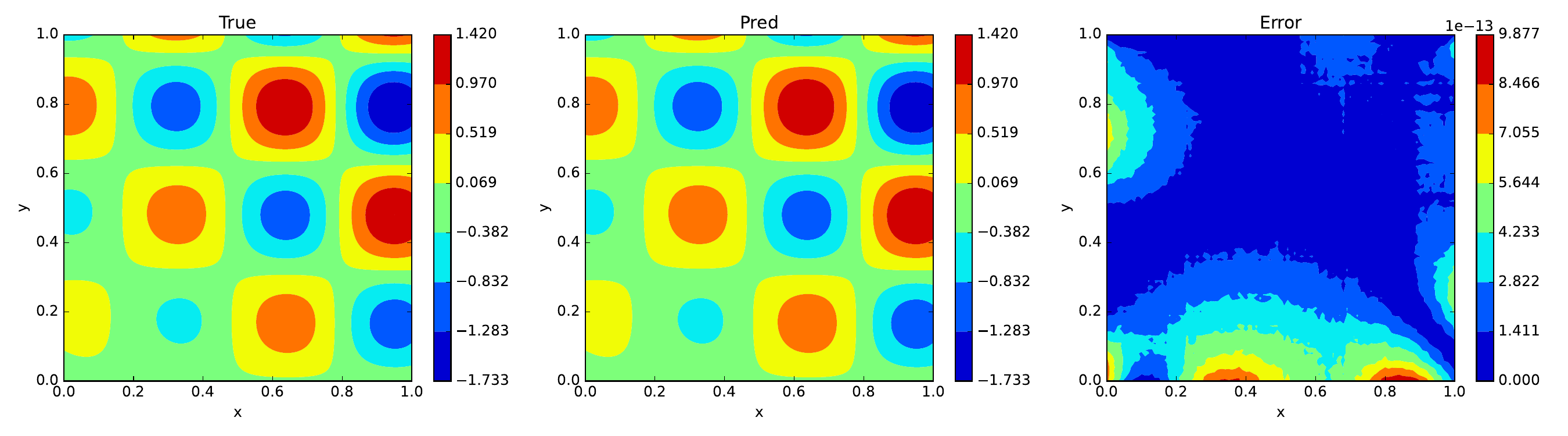}
    \end{minipage}
    \caption{Heat maps illustrate the nonlinear elliptic equation \eqref{eq:NonlinearEllipticEquations}. Left: the exact solution; Middle: the prediction solution of MS-SFNN; Right: the absolute error between them.}
    \label{fig:NonlinearEllipticEquations}
\end{figure}

Table \ref{tab:NonlinearEllipticEquations} reports the $L_{\infty}$ and $L_2$ errors of PINN, SV-SNN, and MS-SFNN for the nonlinear elliptic problem \eqref{eq:NonlinearEllipticEquations}. PINN yields errors around $10^{-1}$, SV-SNN around $10^{-3}$, whereas MS-SFNN reduces both errors to the order of $10^{-13}$, demonstrating exceptional accuracy. Figure \ref{fig:NonlinearEllipticEquations} displays heat maps of the exact solution, the MS-SFNN approximation, and the absolute error, confirming its high fidelity across the domain.

\begin{table}[htp]
    \begin{center}
        \caption{Nonlinear elliptic equations \eqref{eq:NonlinearEllipticEquations}: Performance comparison of PINN, SV-SNN and MS-SFNN. The $L_{\infty}$ errors and $L_{2}$ errors for each model configuration are presented.}
        \begin{tabular}{ccccccc}
            \hline\noalign{\smallskip}
            Method & $e_{L_{\infty}}$ & $e_{L_{2}}$ \\
            \hline
            \text{PINN}\cite{PINN}    & 5.91E-01 & 3.40E-01 \\
            \text{SV-SNN}\cite{SV-SNN}   & 7.21E-03 & 4.05E-03 \\
            \text{MS-SFNN}  & 9.88E-13 & 4.51E-13 \\
            \hline
        \end{tabular}
        \label{tab:NonlinearEllipticEquations}
    \end{center}
\end{table}

\subsection{Double-Cylinder Steady Navier-Stokes Equations}
To further verify the robustness of MS-SFNN on complex geometries and multi-obstacle flow problems, we consider the steady incompressible Navier–Stokes equations in a circular domain with two cylindrical obstacles. The computational domain is $\Omega={(x,y): x^2+y^2 \le 3.0^2} \backslash (\Omega_1 \cup \Omega_2)$, where $\Omega_1$ and $\Omega_2$ are disks of radius $0.3$ centered at $(-1.0, 0.5)$ and $(1.0, -0.5)$, respectively. The governing equations are
\begin{equation}
    \label{eq:DoubleCylinderSteadyNavierStokesEquation}
    \begin{array}{r@{}l}
        \left\{
        \begin{aligned}
            u u_x + v u_y + p_x - \mu (u_xx + u_yy) & = S_1, &  &  \mbox{in} \enspace \Omega, \\
            u v_x + v v_y + p_y - \mu (v_xx + v_yy) & = S_2, &  &  \mbox{in} \enspace \Omega, \\
            u_x + v_y & = 0, &  &     \mbox{in} \enspace \Omega,  \\
            u             & = g_1,        &  & \mbox{on} \enspace \partial \Omega, \\
            v             & = g_2,        &  & \mbox{on} \enspace \partial \Omega, \\
            p             & = g_3,        &  & \mbox{on} \enspace \partial \Omega, \\
        \end{aligned}
        \right.
    \end{array}
\end{equation}
with 
dynamic viscosity $\mu=1.0$. The source terms $S_1$, $S_2$ and the boundary data $g_1, g_2, g_3$ are derived from exact solutions. The exact solution of velocity field is prescribed as $$u = \frac{1}{2} \sin(x+y) + \sin(2x)\cos(2y), \quad v = -\frac{1}{2} \sin(x+y) - \cos(2x)\sin(2y);$$ the exact solution of pressure $p$ is examined in two cases:
\begin{itemize}
    \item Case 1: $p=0$
    \item Case 2: $p = \sin(3x-y) + \sin(x-3y)$
\end{itemize}

For this nonlinear problem, training points are drawn from a uniform $101 \times 101$ grid; points inside or on the two cylindrical obstacles are discarded, and the remaining points serve as collocation points for the PDE and the outer boundary. Each cylinder boundary is discretized with 
$100$ uniformly distributed points using a polar parametrization. 
Since MS-SFNN requires two scaling factors per field, six factors are assigned to the three unknowns $(u,v,p)$: $\rho_{u,1}=\rho_{u,2}=2$, $\rho_{v,1}=\rho_{v,2}=2$,  $\rho_{p,1}=\rho_{p,2}=3$. The nonlinear system is linearized by freezing convective terms with the previous iterate, details are omitted for brevity.

\begin{figure}[htbp]
    \centering
    \begin{minipage}{1.0\linewidth}
        \includegraphics[width=1\textwidth]{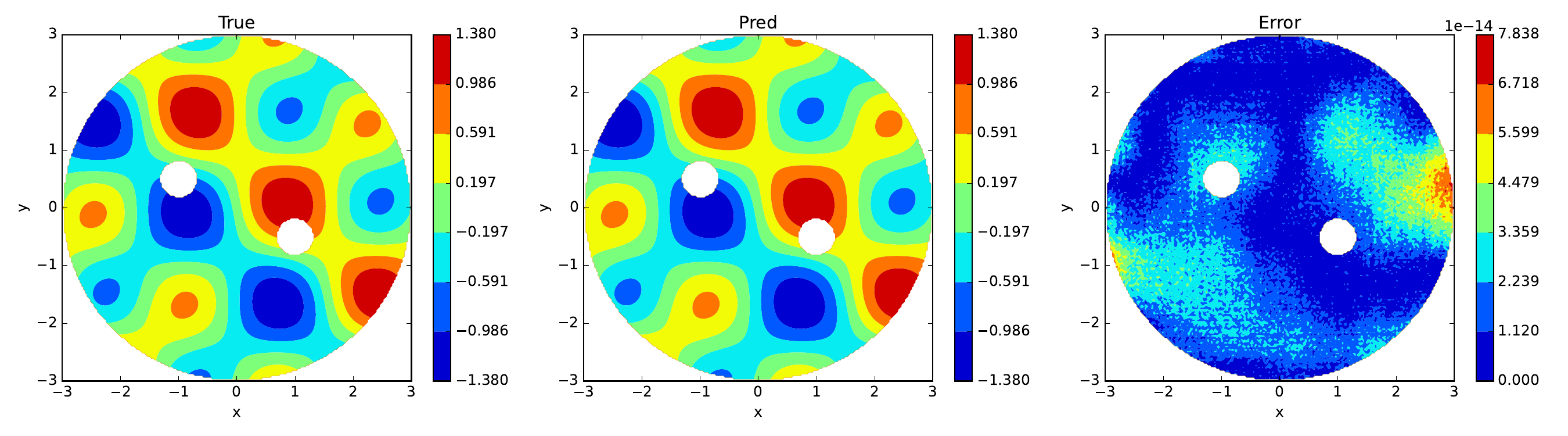}
    \end{minipage}
    \caption{Heat maps illustrate the $u$ component of the double-cylinder steady Navier-Stokes equations \eqref{eq:DoubleCylinderSteadyNavierStokesEquation} in Case $1$. Left: the exact solution; Middle: the prediction solution of MS-SFNN; Right: the absolute error between them.}
    \label{fig:Double-CylinderSteadyNS_u_True}
\end{figure}

\begin{figure}[htbp]
    \centering
    \begin{minipage}{1.0\linewidth}
        \includegraphics[width=1\textwidth]{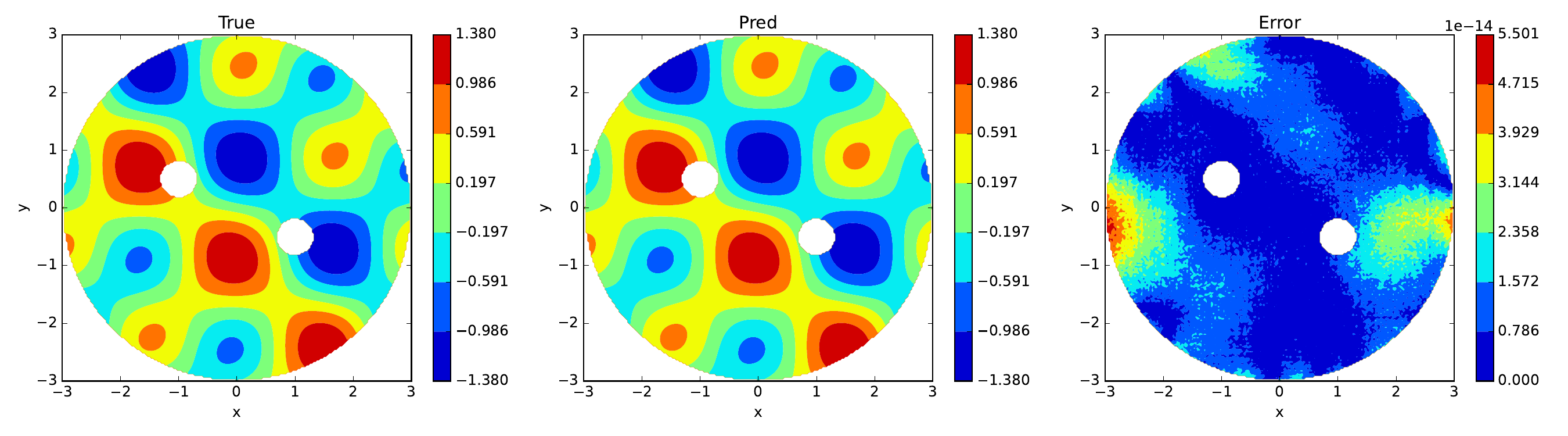}
    \end{minipage}
    \caption{Heat maps illustrate the $v$ component of the double-cylinder steady Navier-Stokes equations \eqref{eq:DoubleCylinderSteadyNavierStokesEquation} in Case $1$. Left: the exact solution; Middle: the prediction solution of MS-SFNN; Right: the absolute error between them.}
    \label{fig:Double-CylinderSteadyNS_v_True}
\end{figure}

The errors for Case 1 ($p=0$) are reported in Table \ref{tab:DoubleCylinderSteadyNavierStokesEquation_True}. For PINN and SV-SNN, the $L_2$ errors of $u,v$ are listed; for MS-SFNN, both $L_\infty$ and $L_2$ errors of $u,v$ and $p$ are given. MS-SFNN achieves errors orders of magnitude smaller than those of PINN and SV-SNN. Figures~\ref{fig:Double-CylinderSteadyNS_u_True} and~\ref{fig:Double-CylinderSteadyNS_v_True} show heat maps of the velocity components $u$ and $v$ (exact, MS-SFNN prediction, and absolute error). The excellent visual agreement confirms that MS-SFNN resolves this nonlinear problem with high accuracy and robustness.

\begin{table}[htp]
    \begin{center}
        \caption{Double-cylinder steady Navier-Stokes equations \eqref{eq:DoubleCylinderSteadyNavierStokesEquation} in Case $1$: Performance comparison of PINN, SV-SNN and MS-SFNN. The $L_{\infty}$ errors and $L_{2}$ errors for each model configuration are presented.}
        \begin{tabular}{ccccccc}
            \hline\noalign{\smallskip}
            \multirow{2}{*}{Method} & \multicolumn{2}{c}{$u$} & \multicolumn{2}{c}{$v$} & \multicolumn{1}{c}{$p$}  \\
            & $e_{L_{\infty}}$ & $e_{L_{2}}$ & $e_{L_{\infty}}$ & $e_{L_{2}}$ & $e_{L_{\infty}}$    \\
            \hline
            \text{PINN}\cite{PINN}    & - & 9.45E-03 & - & 1.50E-02 & -  \\
            \text{SV-SNN}\cite{SV-SNN}   & - & 5.68E-04 & - & 4.06E-04 & -  \\
            \text{MS-SFNN}  & 7.83E-14 & 3.40E-14 & 5.50E-14 & 2.28E-14 & 6.33E-14  \\
            \hline
        \end{tabular}
        \label{tab:DoubleCylinderSteadyNavierStokesEquation_True}
    \end{center}
\end{table}

\begin{figure}[htbp]
    \centering
    \begin{minipage}{1.0\linewidth}
        \includegraphics[width=1\textwidth]{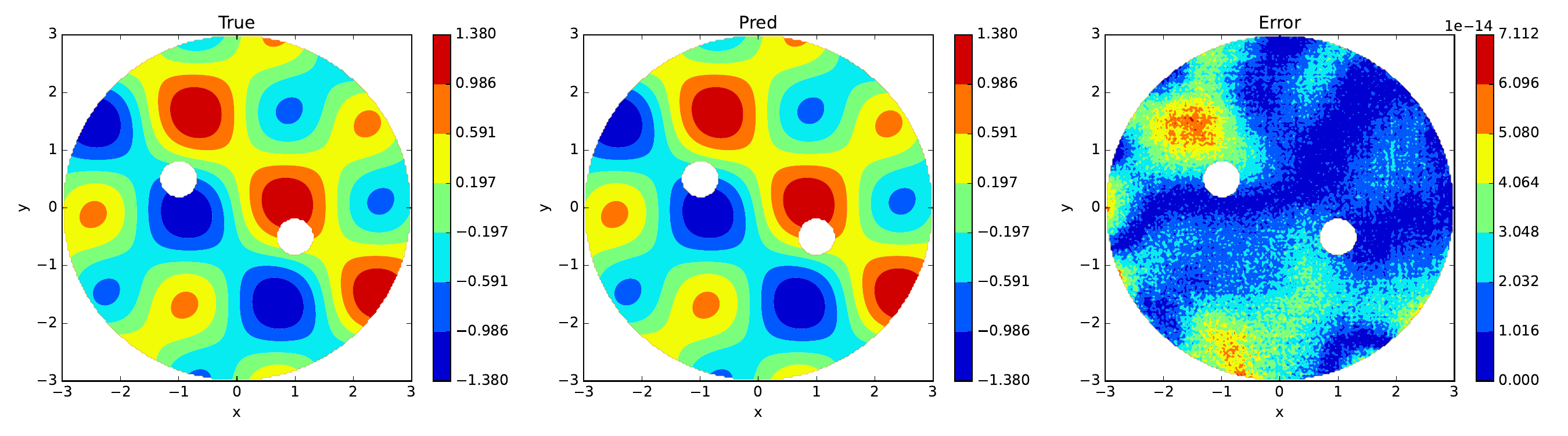}
    \end{minipage}
    \caption{Heat maps illustrate the $u$ component of the double-cylinder steady Navier-Stokes equations \eqref{eq:DoubleCylinderSteadyNavierStokesEquation} in Case $2$. Left: the exact solution; Middle: the prediction solution of MS-SFNN; Right: the absolute error between them.}
    \label{fig:Double-CylinderSteadyNS_u_False}
\end{figure}

\begin{figure}[htbp]
    \centering
    \begin{minipage}{1.0\linewidth}
        \includegraphics[width=1\textwidth]{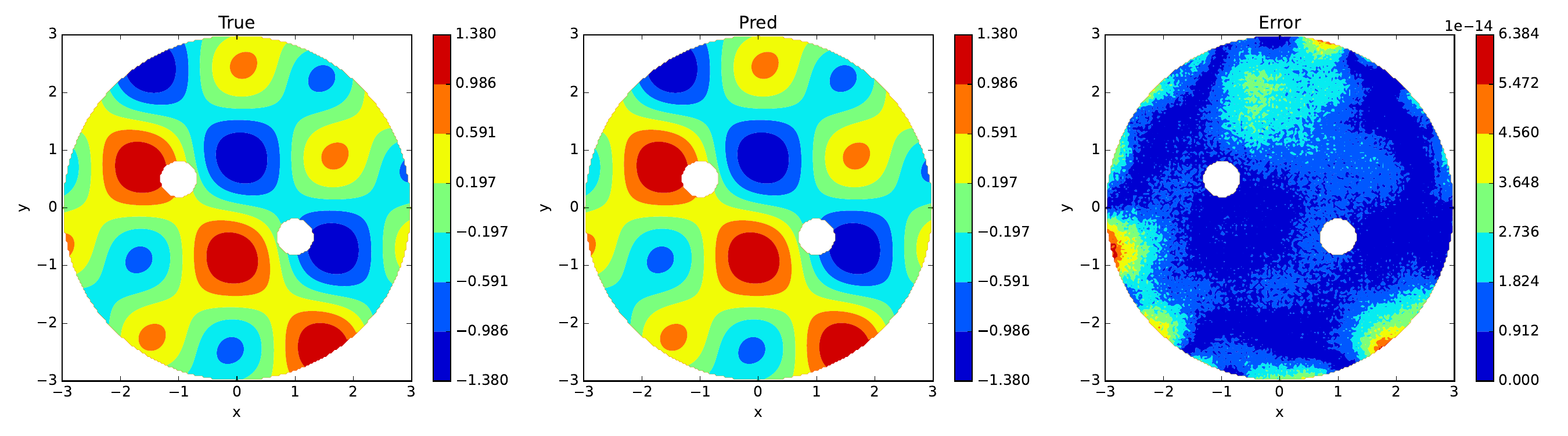}
    \end{minipage}
    \caption{Heat maps illustrate the $v$ component of the double-cylinder steady Navier-Stokes equations \eqref{eq:DoubleCylinderSteadyNavierStokesEquation} in Case $2$. Left: the exact solution; Middle: the prediction solution of MS-SFNN; Right: the absolute error between them.}
    \label{fig:Double-CylinderSteadyNS_v_False}
\end{figure}

\begin{figure}[htbp]
    \centering
    \begin{minipage}{1.0\linewidth}
        \includegraphics[width=1\textwidth]{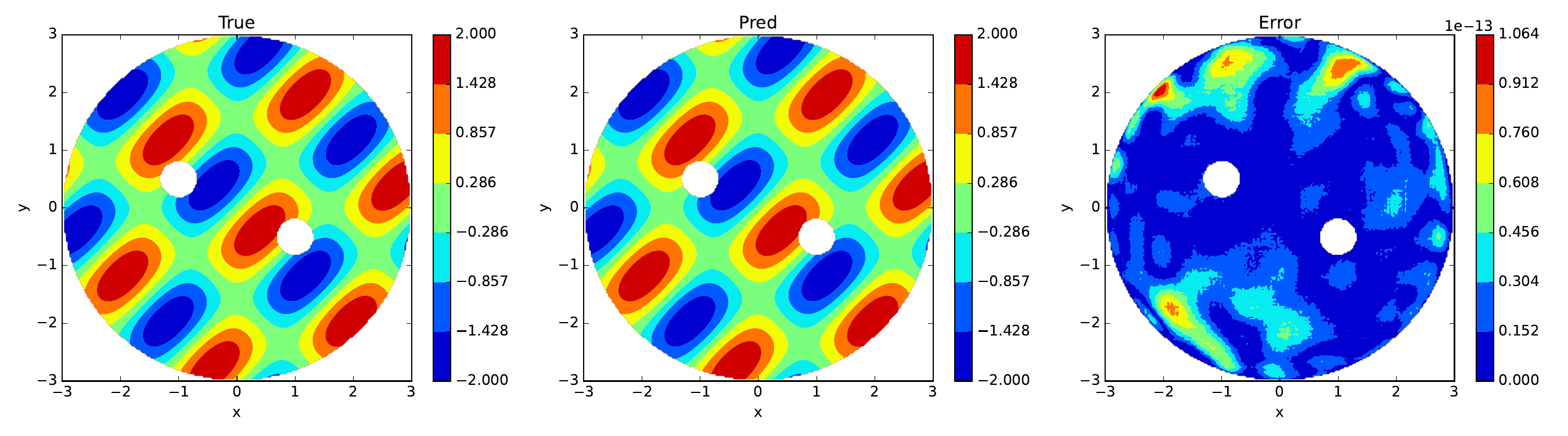}
    \end{minipage}
    \caption{Heat maps illustrate the $p$ component of the double-cylinder steady Navier-Stokes equations \eqref{eq:DoubleCylinderSteadyNavierStokesEquation} in case $2$. Left: the exact solution; Middle: the prediction solution of MS-SFNN; Right: the absolute error between them.}
    \label{fig:Double-CylinderSteadyNS_p_False}
\end{figure}

The $L_{\infty}$ and $L_2$ errors of $u,v$ and $p$
obtained by MS-SFNN for Case 2 are reported in Table \ref{tab:DoubleCylinderSteadyNavierStokesEquation_False}, where the non-zero pressure field increases the problem complexity. MS-SFNN maintains high accuracy, as confirmed by Figures \ref{fig:Double-CylinderSteadyNS_u_False}--\ref{fig:Double-CylinderSteadyNS_p_False}, which display the exact solution, the MS-SFNN approximation, and the absolute error for each field. The minimal absolute errors demonstrate the method's robustness and effectiveness on this nonlinear problem.

\begin{table}[htp]
    \begin{center}
        \caption{Double-cylinder steady Navier-Stokes equations \eqref{eq:DoubleCylinderSteadyNavierStokesEquation} in Case $2$: Performance of MS-SFNN. The $L_{\infty}$ errors and $L_{2}$ errors for each model configuration are presented.}
        \begin{tabular}{ccccccc}
            \hline\noalign{\smallskip}
            \multirow{2}{*}{Method} & \multicolumn{2}{c}{$u$} & \multicolumn{2}{c}{$v$} & \multicolumn{2}{c}{$p$}  \\
            & $e_{L_{\infty}}$ & $e_{L_{2}}$ & $e_{L_{\infty}}$ & $e_{L_{2}}$ & $e_{L_{\infty}}$ & $e_{L_{2}}$     \\
            \hline
            \text{MS-SFNN}  & 7.11E-14 & 3.84E-14 & 6.38E-14 & 2.53E-14 & 1.06E-14 & 2.33E-14  \\
            \hline
        \end{tabular}
        \label{tab:DoubleCylinderSteadyNavierStokesEquation_False}
    \end{center}
\end{table}

\subsection{Car Cabin Acoustic Simulation}
We simulate the interior acoustics of a car cabin using the three-dimensional Helmholtz equation with mixed boundary conditions.
The cabin is modeled as the cuboid $[-1.5, 1.5] \times [-2, 0] \times [-1, 1]$; exterior components are omitted. The governing equation is \eqref{eq:Helmholtz_Equation}, subject to
\begin{equation}
    \label{eq:CarCabin_Equation}
    \begin{array}{r@{}l}
        \left\{
        \begin{aligned}
            u_n             & = g_1,        &  & \mbox{on} \enspace \Gamma_1, \\
            u_n + \mathrm{i} u            & = g_2,        &  & \mbox{on} \enspace \Gamma_2, 
        \end{aligned}
        \right.
    \end{array}
\end{equation}
where $\Gamma_1$ is the bottom surface and $\Gamma_2$ comprises the remaining boundaries.

\begin{figure}[htbp]
    \centering
    \begin{minipage}{1.0\linewidth}
        \includegraphics[width=1\textwidth]{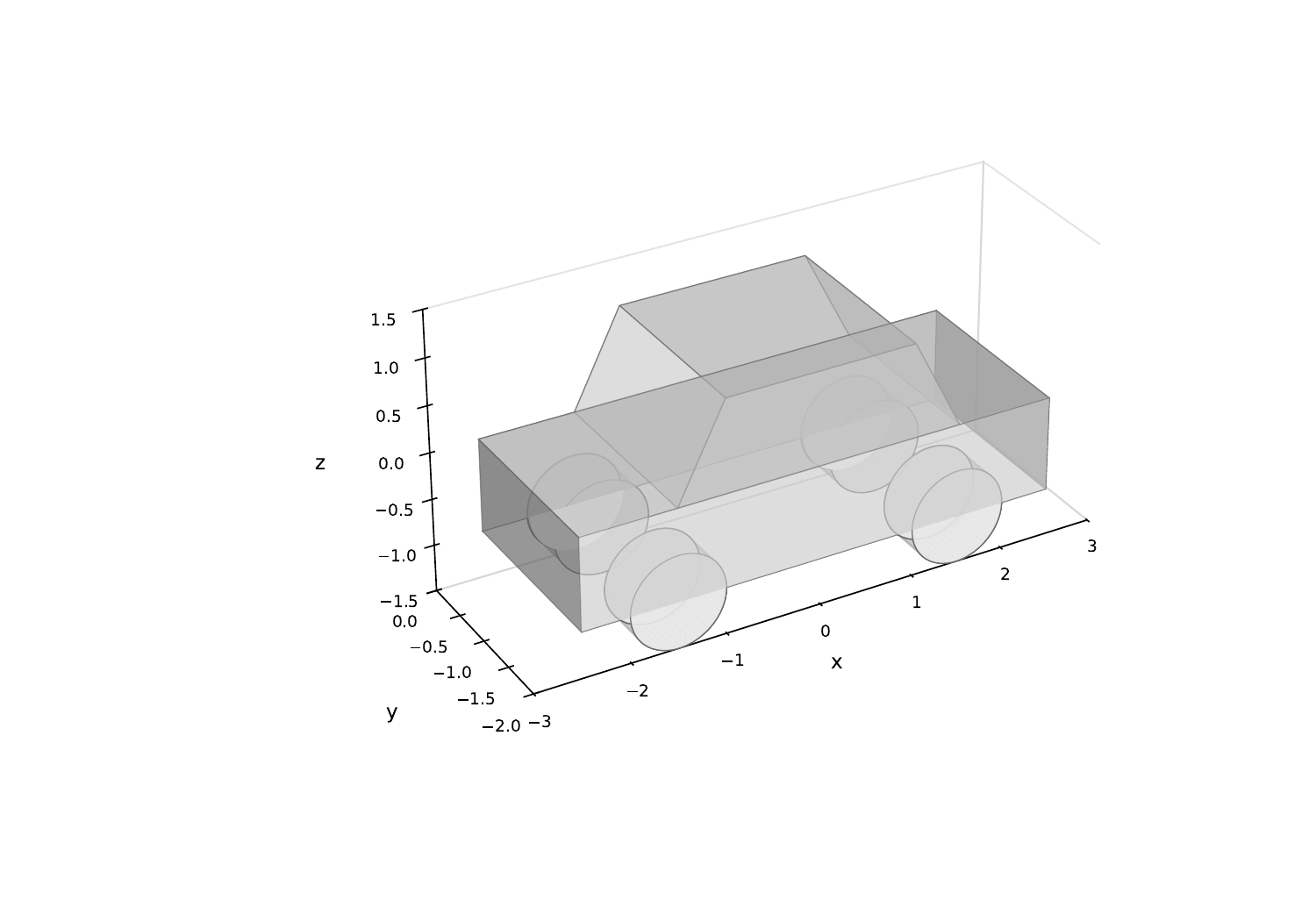}
    \end{minipage}
    \caption{Three-dimensional car model.}
    \label{fig:carcabin}
\end{figure}

Six acoustic sources (all outside $\Omega$) are placed at  $\boldsymbol{\hat{x}}_1=(0, -1, 2)$, $\boldsymbol{\hat{x}}_2=(2,-1,0)$, $\boldsymbol{\hat{x}}_3=(0,1,0)$, $\boldsymbol{\hat{x}}_4=(0, -1, -2)$, $\boldsymbol{\hat{x}}_5=(-2, -1, 0)$, and $\boldsymbol{\hat{x}}_6=(0, -3, 0)$. The exact solution is $u(\boldsymbol{x})=\sum_{i=1}^6 H(\boldsymbol{\hat{x}}_i, \boldsymbol{x})$ with  $H(\boldsymbol{x}, \boldsymbol{y})=\frac{1}{4\pi} \frac{\exp(\mathrm{i} k \lvert \boldsymbol{x} - \boldsymbol{y} \rvert)}{ \lvert \boldsymbol{x} - \boldsymbol{y} \rvert}$, where  $k=\frac{2 \pi v_{freq}}{340}$ and $v_{freq}=1000$. The boundary data $g_1$, $g_2$ are derived from this exact solution. Figure \ref{fig:carcabin} illustrates the car model.

Training points are generated on a uniform $100 \times 100 \times 100$ grid, from which only interior points of the cabin are retained. On each rectangular boundary face, a 
$100 \times 100$ uniform grid is created and points not lying on the face are discarded; on the two curved surfaces, a $100 \times 100$ grid is sampled using polar coordinates. Since the problem is complex-valued, we decompose the approximate solution into real and imaginary parts:
\begin{equation}
    \label{eq:carcabin_uv}
    \begin{array}{r@{}l}
        \left\{
        \begin{aligned}
    \text{Re}\ u_M(\boldsymbol{x}) &= \sum_{i=1}^M w_{\text{real},i} \prod_{j=1}^d \cos(\rho_{\text{real},j}(w_{ji} x_j + b_{ij})), \\
    \text{Im}\ u_M(\boldsymbol{x}) &= \sum_{i=1}^M w_{\text{imag},i} \prod_{j=1}^d \cos(\rho_{\text{imag},j}(w_{ji} x_j + b_{ij})),
        \end{aligned}
        \right.
    \end{array}
\end{equation}
where all scaling factors are set to the wavenumber, i.e., $\rho_{\text{real},j} = \rho_{\text{imag},j} = k. $ The weights $w_{ji}$and biases $b_{ij}$ are randomly initialized and kept fixed.

\begin{figure}[htbp]
    \centering
    \begin{minipage}{1.0\linewidth}
        \includegraphics[width=1\textwidth]{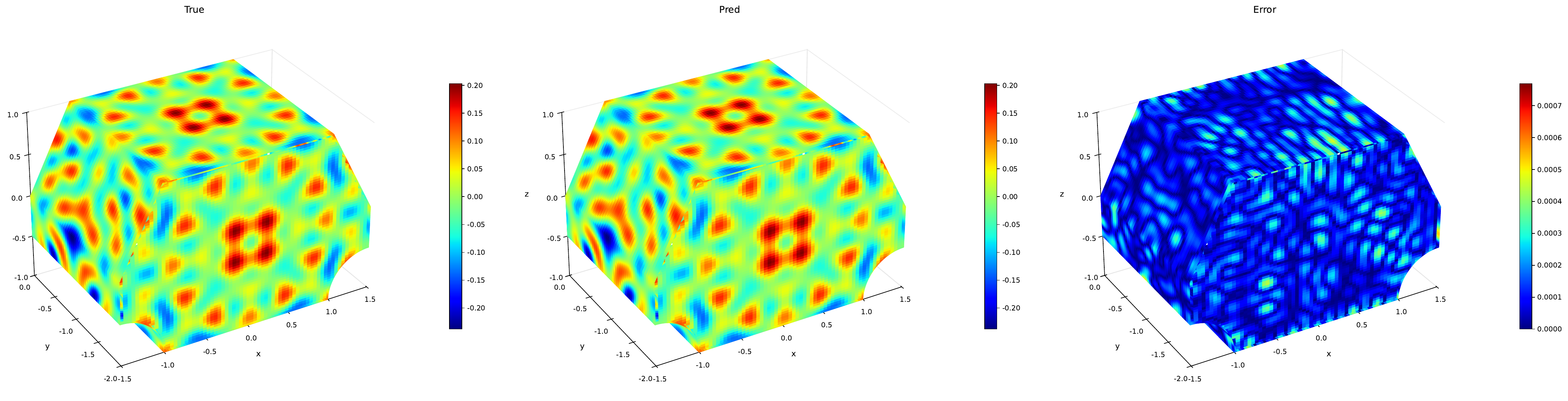}
    \end{minipage}
    \caption{Heat maps illustrate the three-dimensional car cabin acoustic problem. 
    Left: the exact solution; Middle: the prediction solution of MS-SFNN; Right: the absolute error between them.}
    \label{fig:carcabin_error}
\end{figure}

Table \ref{tab:carcabin_error} reports the $L_{\infty}$ and $L_2$ errors of PINN and MS-SFNN. For both the real and imaginary parts, PINN’s errors are substantially larger than those of MS-SFNN, indicating that PINN struggles with this problem while MS-SFNN achieves much higher accuracy. Figure \ref{fig:carcabin_error} displays 3D heat maps of the exact solution, the MS-SFNN prediction, and the absolute error; MS-SFNN accurately reproduces the exact solution.

\begin{table}[htp]
    \begin{center}
        \caption{Car cabin acoustic problem: Performance comparison of PINN and MS-SFNN. The $L_{\infty}$ errors and $L_{2}$ errors for each model configuration are presented.}
        \begin{tabular}{ccccc}
            \hline\noalign{\smallskip}
            \multirow{2}{*}{Method} & \multicolumn{2}{c}{Re $u$} & \multicolumn{2}{c}{Im $u$} \\
            & $e_{L_{\infty}}$ & $e_{L_{2}}$ & $e_{L_{\infty}}$ & $e_{L_{2}}$ \\
            \hline
            \text{PINN}\cite{PINN} & 2.03E-01 & 9.94E-01 & 1.30E-01 & 1.02E+00 \\
            \text{MS-SFNN}  & 5.81E-04 &1.07E-03 & 5.08E-04 & 8.55E-04 \\
            \hline
        \end{tabular}
        \label{tab:carcabin_error}
    \end{center}
\end{table}

\subsection{Effects of scale factors $\rho_j$}
To study the effect of scaling factors $\rho_j (j = 1,...,d)$, we solve the two-dimensional Helmholtz equation \eqref{eq:Helmholtz_Equation}--\eqref{eq:Helmholtz_EquationDich} with MS-SFNN, performing a grid search  over $\rho_1, \rho_2 \in [1,200]$. 
The exact solution is taken as $u=\sin(k_1 x)\sin(k_2 y)$ with $k =48\pi$, in two cases:
\begin{itemize}
    \item Case 1: $k_1=k_2 = k$; 
    \item Case 2: $k_1=k$ and $k_2= \frac{k}{2}$. 
\end{itemize}

\begin{figure}[htbp]
    \centering
    \begin{minipage}{0.48\linewidth}
        \includegraphics[width=1\textwidth]{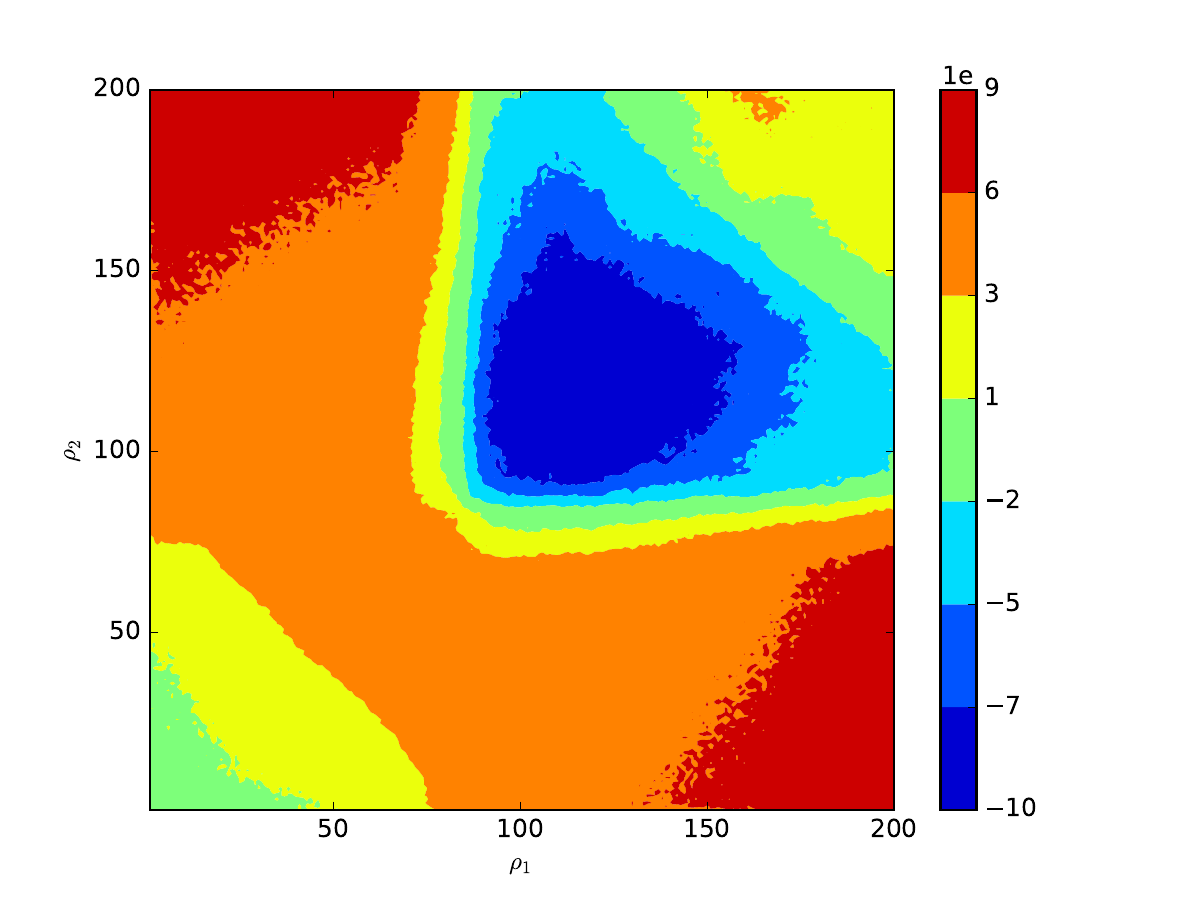}
    \end{minipage}
    \begin{minipage}{0.48\linewidth}
        \includegraphics[width=1\textwidth]{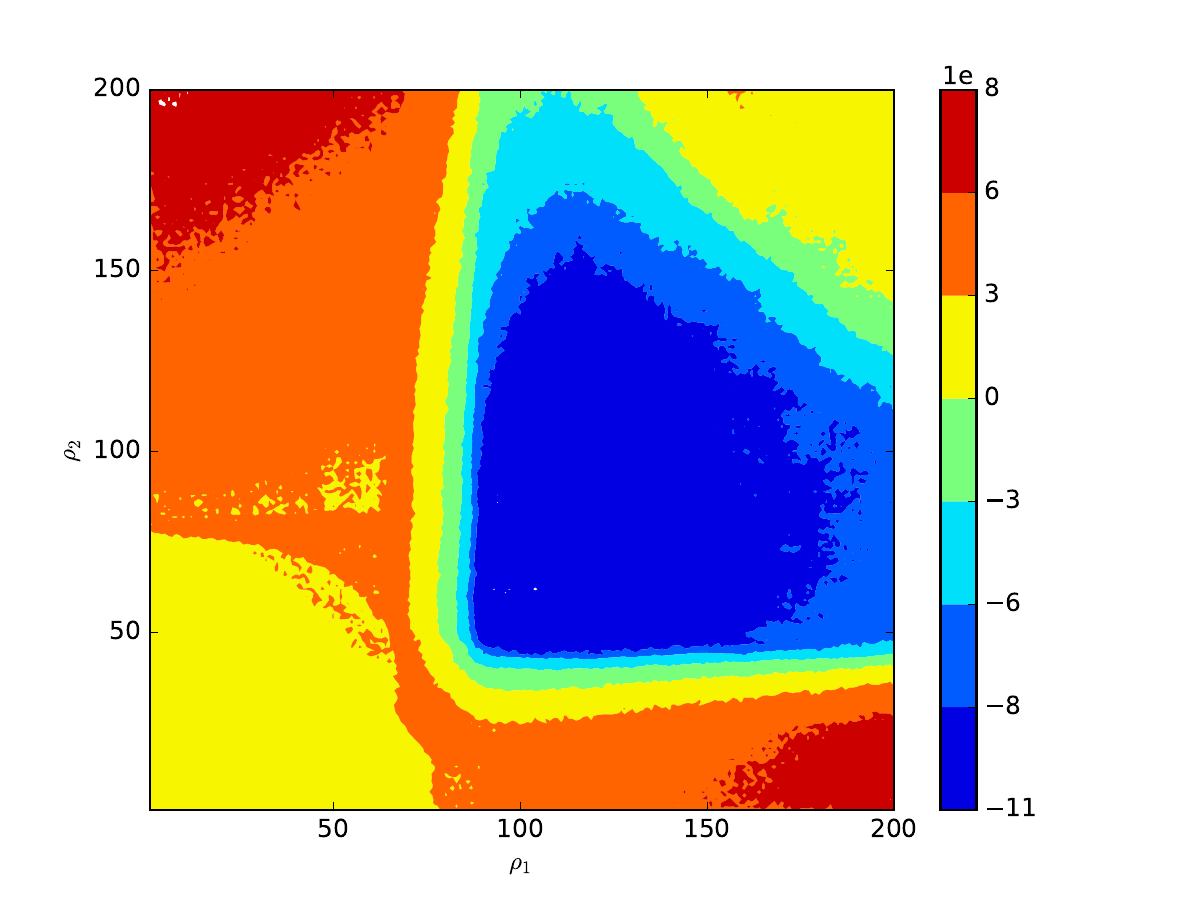}
    \end{minipage}
    \caption{ 
    Heat maps of the $L_{\infty}$
  errors of $u$ as a function of $\rho_1$ and $\rho_2$ (each ranging from $1$ to $200$) on a logarithmic color scale. Left: Case $1$; right: Case $2$. }
    \label{fig:rhos_infty}
\end{figure}

In Figure \ref{fig:rhos_infty}, we present a heat map (logarithmic color scale) illustrating how the $L_{\infty}$ error of the approximate solution of MS-SFNN varies with respect to the scaling factors $\rho_1$ and $\rho_2$ for both cases: the left panel corresponds to Case 1, and the right panel to Case 2.

In Case 1 ($k_1 = k_2$), the $x$ and $y$ directions influence the solution equally. The corresponding heat map (Figure~\ref{fig:rhos_infty}, left) is nearly symmetric about the diagonal $\rho_1 = \rho_2$, indicating that the error depends primarily on the deviation from this symmetry axis. Hence, setting $\rho_1=\rho_2$ is both natural and optimal. In Case $2$ ($k_1 \neq k_2$), the directions have different scales of influence. The heat map (Figure~\ref{fig:rhos_infty}, right) reflects this asymmetry: the optimal $\rho_1$ is larger than the optimal 
$\rho_2$, consistent with $k_1 > k_2$. This demonstrates that distinct scaling factors are necessary to match the directional sensitivity of the solution.

These observations highlight a critical practical insight: without carefully selection of $(\rho_1, \rho_2)$, MS-SFNN may fail to recover a physically meaningful solution, yielding errors so large that the result is numerically unusable. Proper tuning of the scaling factors is therefore not just beneficial—it is essential for the method to succeed.


\subsection{Comparison Experiments with Existing Methods}


To precisely assess MS-SFNN for high-frequency regimes, we compare it with XPINN \cite{XPINN}, FBPINN \cite{FBPINN}, FourierPINN \cite{FourierPINN}, BsPINN \cite{BsPINN}, and SV-SNN \cite{SV-SNN} on the two-dimensional Helmholtz equation in Section \ref{sec4.1} with $k=24\pi$, averaging over 
$10$ random initializations (Table \ref{tab:ComparisonExistingMethods}). While the competing methods yield average $L_2$ errors of order  $10^{-1}$, MS-SFNN achieves errors of order $10^{-12}$---over ten orders of magnitude more accurate than the best baseline---with negligible standard deviation. This stark gap demonstrates that MS-SFNN not only provides unprecedented accuracy but also exceptional reliability and robustness, effectively overcoming the spectral bias and optimization difficulties that hinder conventional PINNs in oscillatory settings.

\begin{table}[htp]
    \begin{center}
        \caption{Two dimensional Helmholtz equations \eqref{eq:Helmholtz_Equation}--\eqref{eq:Helmholtz_EquationDich} with $k=24\pi$: Performance comparison of XPINN, FBPINN, FourierPINN, BsPINN, SV-SNN and MS-SFNN. The average value and standard deviation of $L_{2}$ errors for each model configuration are presented.}
        \begin{tabular}{ccccccc}
            \hline\noalign{\smallskip}
            Method & average $e_{L_{2}}$ & std $e_{L_{2}}$ \\
            \hline
            \text{XPINN}\cite{XPINN}    & 9.44E-01 & 1.62E-01 \\
            \text{FBPINN}\cite{FBPINN}  & 6.79E-01 & 3.28E-01 \\
            \text{FourierPINN}\cite{FourierPINN}  &3.06E-01 & 8.17E-02 \\
            \text{BsPINN}\cite{BsPINN}    & 1.74E-01 & 5.69E-01 \\
            \text{SV-SNN}\cite{SV-SNN}   & 1.27E-02 & 2.05E-02 \\
            \text{MS-SFNN}  & 2.61E-12 & 8.44E-13 \\
            \hline
        \end{tabular}
        \label{tab:ComparisonExistingMethods}
    \end{center}
\end{table}

\section{Conclusions}
\label{sec:conclusions}
In this work, we propose Multi-Scale Separable Fourier Neural Networks (MS-SFNN) for high-frequency PDEs. The core idea is a separable representation: for a $d$-dimensional input, $d$ independent subnetworks each process one coordinate, and their outputs are multiplied element-wise to form separable basis functions. The PDE solution is expressed as a linear combination of these bases, with coefficients determined by least squares. The weights and biases of each subnetwork are randomly initialized (uniform, variance 1) and then kept fixed. To overcome the limited expressivity of fixed random features, we multiply each subnetwork’s weights and biases by a scaling factor, which directly controls the frequency content of the basis functions; each subnetwork has its own scaling factor, so the overall architecture contains several such factors. A cosine activation is used to explicitly incorporate Fourier features, endowing MS-SFNN with strong oscillatory representation capabilities well suited for high-frequency problems.

High-frequency and three-dimensional problems require a large number of collocation points to resolve fine-scale oscillations, dramatically increasing memory demands and often exceeding GPU capacity. To address this, we introduce two key innovations. First, instead of AD, we derive and employ closed-form analytical expressions for all spatial and temporal derivatives of the basis functions, eliminating computational graphs and drastically reducing memory and evaluation time. Second, we use a memory-efficient batched QR decomposition algorithm that processes collocation points in small chunks and incrementally builds an orthogonal representation, avoiding the need to store the full basis matrix. Together, these strategies significantly reduce peak GPU memory and enable MS-SFNN to scale effectively to high-frequency and three-dimensional regimes while retaining high accuracy.

Despite its strong performance, MS-SFNN has notable limitations that motivate future work. Accuracy depends critically on the scaling factors, and poorly chosen values can degrade the solution by orders of magnitude. A brute-force grid search is computationally prohibitive in high dimensions, so an efficient and robust method for automatically determining near-optimal scaling factors is needed. Moreover, for very high frequencies or three-dimensional high-frequency problems, the method exhibits a noticeable loss of precision, indicating that the current basis construction may not fully capture the most challenging spectral components. Future research should therefore focus on designing enhanced basis architectures capable of maintaining high accuracy in ultra-high-frequency and three-dimensional settings.

\section*{Acknowledgment}
This research is partially supported by the National Natural Science Foundation of China (No.12371434, No.U25A20200), and the National Key R \& D Program of China (No.2022YFE03040002).

\section*{Data Availability Statement}
The data that support the findings of this study are available from the corresponding author upon reasonable request.

\bibliographystyle{unsrt}
\bibliography{./ref}
\end{document}